\pdfoutput=1
\documentclass[11pt]{article}

\usepackage{emnlp2023}

\usepackage{times}
\usepackage{latexsym}

\usepackage[T1]{fontenc}
\usepackage[utf8]{inputenc}
\usepackage[subtle]{savetrees}
\usepackage{multirow}
\usepackage{hyperref}
\usepackage{booktabs} % for professional tables
\usepackage{tabularx}
\usepackage{graphicx}
\usepackage{subcaption}
\usepackage[htt]{hyphenat}
\usepackage{sidecap}
\usepackage[T1]{fontenc}
\usepackage{ragged2e}
\usepackage{siunitx}
\usepackage{adjustbox}
\usepackage{amsmath,amsfonts,bm}
\usepackage{mathtools}
\usepackage{microtype}
\usepackage{todonotes}

\title{Smaller Language Models are Better  \\Black-box Machine-Generated Text Detectors}

\author{Fatemehsadat Mireshghallah\textsuperscript{\rm 1, \rm 2}, Justus Mattern\textsuperscript{\rm 3}, Sicun Gao\textsuperscript{\rm 1}\\
   \textbf{Reza Shokri}\textsuperscript{\rm 4},    \textbf{Taylor Berg-Kirkpatrick}\textsuperscript{\rm 1} \\
    \textsuperscript{\rm 1} University of California San Diego,
        \textsuperscript{\rm 2} University of Washington,
    \textsuperscript{\rm 3} RWTH Aachen\\
    \textsuperscript{\rm 4} National University of Singapore \\
    \texttt{[fatemeh, sicung,tberg]@ucsd.edu},\\ \texttt{ justus.mattern@rwth-aachen.de,reza@comp.nus.edu.sg}
  }

\begin{document}
\maketitle
\begin{abstract}

% Recent zero-shot machine generated text method, DetectGPT~\cite{mitchell2023detectgpt}, relies on the curvature/spiky-ness of the loss function for a given model,  to distinguish  text generated by it from human-written text. 
% %
% They do so by generating `neighboring perturbations' for each target sequence and comparing the loss of the target with it s neighbors.
% %
% They observe that some models are better at detecting the text generated by other models, but they do not discuss the patterns or possible reasons for some models being superior.
%that can  produce convincing utterances very similar to those written by humans, d
%%%%%%%%%%%%%%%%%%
% With the advent of fluent generative language models, distinguishing whether a piece of text is machine-generated or human-written has become more challenging and more important, as such models could be used to spread misinformation and to mimic certain authors and figures.
% %
% To this end, there have been a slew of methods proposed to detect machine-generated text, most of which need access to the loss of the target model or need the ability to sample from the target.
% %
% %
% One such black-box detection method relies on the observation that generated text is locally optimal under the likelihood function of the generator, while human-written text is not.
%%%%%%%%%%%

As large language models are becoming more ubiquitous and embedded in different user-facing services, it is important to be able to distinguish between human-written  and machine-generated text, to verify the authenticity of news articles, product reviews, etc.
Thus, in this paper we set out to explore \textit{whether it is possible to use one language model to identify machine-generated text produced by another language model, even if the two have different architectures and are trained on different data. Further, if this is possible, which language models make the best general-purpose detectors?}
We find that overall, \textit{smaller and partially-trained models are better universal machine-generated text detectors}: they can more precisely detect text generated from both small and larger models. Interestingly,  we find that whether or not the detector and generator models were trained on the same data or have similar parameter counts is not critically important to the detection success.
For instance the OPT-125M model has an AUC of 0.81 in detecting ChatGPT generations, whereas a larger model from the GPT family, GPTJ-6B, has AUC of 0.45.

\end{abstract}

% \begin{figure}[]
% \centering
%   \label{fig:main_heatmap}
%       \includegraphics[revwidth=\linewidth]{figs/manifold.png}
%      \caption{Loss shape for different models -- My Hypothesis}  
% \end{figure}

\section{Introduction}

\begin{figure*}[t!]
\centering
  \label{fig:main_heatmap_checkpoint}
      \includegraphics[width=0.95\linewidth]{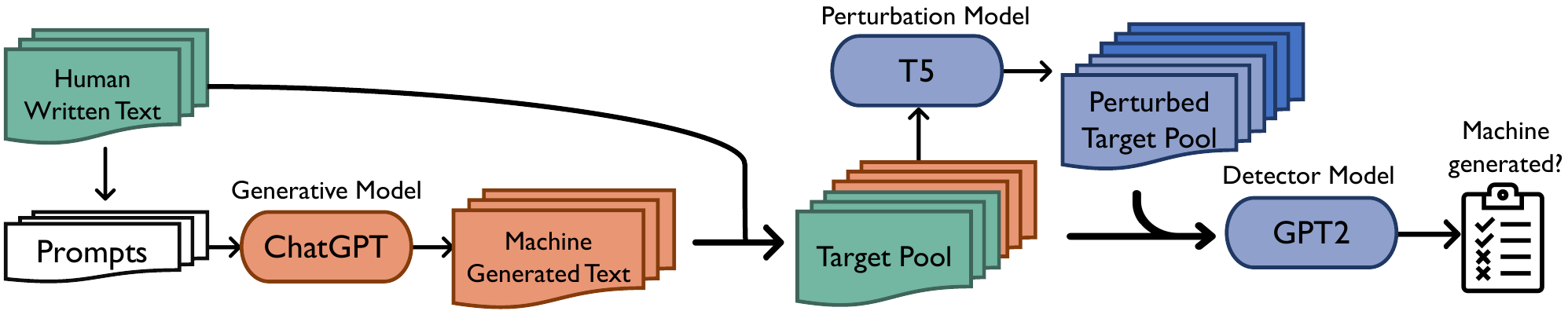}
     \caption{We want to study how models can \textit{cross-detect}, i.e.~distinguish between human-written  text and machine-generated text generated by another model. To this end, we create a \textit{target pool} consisting of both human-written and machine-generated text. We then generate perturbations of each target sequence using a \textit{perturbation model}. We find the likelihood of the target pool and perturbations under a \textit{detector model} in order to estimate the local optimality under the detector model's likelihood. We use the estimate of local optimality to determine if a sequence is machine generated or not.   }  
     \vspace{-1ex }
     \label{fig:methodology}
\end{figure*}

With the rapid improvement in fluency of the text generated by large language models (LLMs), these system are being adopted more and more broadly in a wide range of applications, including chatbots, writing assistants, and summarizers. Generations from these models are shown to have human-like fluency~\cite{liang2022holistic,yuan2022wordcraft}, making it difficult for human readers to differentiate  machine-generated text from human-written text. This can  have significant ramifications, as  such LLM-based tools can be abused for unethical purposes  like  phishing, astroturfing, and generating fake news~\cite{he2023mgtbench}. 
As such, we need to be able to reliably and automatically detect machine generated text.
%, especially in settings where we have no access to the generator model.

%(3) need to be able to reliably and automatically detect machine generated text from black-box, or worse, completely unknown models
%

%There is a risk that AI tools can be abused for unethical purposes like generating fake news, plagiarism, spamming, creating fake product reviews, and manipulating web content for social engineering, which can have negative impacts on society. AI-generated news articles can also have fundamental errors, so it's crucial to differentiate human-written text from AI-generated sequences.
%

Previous work has found that identifying local optima in the likelihood surface of a trained language model allows for detection of training utterances~\cite{mattern2023}, and generations for a given target model~\cite{mitchell2023detectgpt}. Specifically, the approximate measure of local optimality, dubbed curvature, is formed by comparing the loss of a target sequence to the loss of nearby perturbations of the target sequence, under the target model.
The intuition in both prior works is that this measure of curvature is \textit{higher} around both training examples and model generations, compared to unseen human-written text and can therefore be used to determine if a given sequence is part of the training data or not~\cite{mattern2023} or a generation of the target (generator) model or not~\cite{mitchell2023detectgpt}.
%Further, additional prior work has found that the same type of measure can be used to detect *generated* utterances even if they did not appear in train. 
%

In practice, however, we often want to distinguish between machine-generated text and human-written text in situations where we do not know which model could have been used as the generator --- and even if we do know the generator, we might not have access to its likelihood function (e.g. ChatGPT), or access might be behind a paywall (e.g. GPT3). Therefore, in this paper we set out to explore the detection of machine-generated text without knowledge of the generator.
%{whether it is possible to use one language model to identify machine-generated text produced by another language model, even if the two have different architectures and are trained on different data
We do this by exploring whether it is possible to use  the curvature metric measured on one language model (a \textit{detector} model) to identify machine-generated text  generated by another language model (~\textit{the generator}), and under what conditions such cross-detection performs best. We use surrogate detector models, whose likelihood functions we do have access to. Then, we run the curvature test using the surrogate (see Figure~\ref{fig:methodology}) and compare detection power with the same test, but using the true generator's likelihood. 
%
%
%To this end, we study how a recent machine-generated text detection method,  DetectGPT~\cite{mitchell2023detectgpt} can be used for this.
%
%In our paper, however, we want to see how transferable this method is, as in what happens if we use the loss of other generative models, to~\textit{cross-detect} generations from other (potentially unknown/inaccessible) models, and what patterns there are in the performance of the detection power of different models.

%DetectGPT uses the shape and curvature of  loss function of the generative model to infer whether or not a certain sequence is generated.
%
%The intuition is that the loss function should curve around the generations, but not around the human written text.
%
%In our paper, however, we want to see how transferable this method is, as in what happens if we use the loss of other generative models, to~\textit{cross-detect} generations from other (potentially unknown/inaccessible) models, and what patterns there are in the performance of the detection power of different models.
%

We conduct an extensive empirical analysis by experimenting on a slew of models with different sizes (from tens of millions to billions of parameters), architectures (GPTs, OPTs, Pythias) and pre-training data (Webtext and the Pile) and also from different training stages (ranging from the first thousand steps of training to full training-- 143k steps).
%
%on a collection of $23$ models (ranging in size from 70M to 6.7B parameters), using them as detectors for generations of $15$ target models, which include ChatGPT and GPT-3 (davinci).
%
Our main finding is that \textit{cross-detection can come very close to self-detection in terms of distinguishablity}, and that \textit{there are universal cross-detectors with high average distinguishablity} performance, meaning they perform well in terms of detecting generations from a wide-range of models, regardless of the architecture or training data.
More specifically, we find that \textit{smaller models are better universal detectors}. For instance the OPT-125M model  comes within $0.07$  area under the ROC curve of self-detection, on average (see Figure~\ref{fig:main_heatmap_size_diff}). And for models where we don't have self-detection, such as ChatGPT, the AUC of using OPT-125M is $0.81$, whereas OPT 6.7B's AUC is $0.58$. 

We also find that \textit{partially trained models are better detectors} than the fully trained ones, and this gap is bigger for larger models (see Figure~\ref{fig:main_heatmap_checkpoint_diff}).
We then further investigate some possible reasons for this phenomenon by analyzing curvature and log-likelihood of the different models,
and find that larger models are more conservative in terms of the likelihood and curvature they assign to generations from other models. Smaller models, however, assign higher curvature to generations of models their size or larger, therefore they can be used to cross-detect on a broader range of models so the smaller model is the best universal detector.

\section{Methodology}\label{sec:methodology}

Figure~\ref{fig:methodology} shows the methodology of our work, and how we conduct our experiments: For a given \textit{target pool} of sequences, the task is to \textit{determine if each sequence is human-written or machine-generated} by running a curvature (local optimality) test using the likelihood surface of a surrogate \textit{detector model} that is different from the generator model, as our main assumption is that \textit{we have no information about the generator model}.
In the rest of this section we delve deeper into the details of each component in the setup. 

\noindent\textbf{Target pool.} The pool of sequences for which we want to conduct the machine-generated text detection. We form this pool such that there is a 50\%/50\% composition of machine-generated/human-written text.
The machine-generated text is created by prompting the \textit{generator model} with the first $20$ tokens of each human-written sequence.

\noindent\textbf{Generator model.} This model is the generator of the machine-generated utterances we would like to distinguish from human-written utterances.
We do not always have full access to this model, or even know what model it is. 
This scenario is what we are actually interested in: we want to know \textit{how we detect text generated by unknown models.}

%\paragraph{Target Model.}
%Target model is the model the generations of which we are trying to distinguish from human written text.
%

\noindent\textbf{Detector model.}
 This model is used as a \textit{surrogate} for the target model, to help us detect generations when using the curvature test. 
The pool of sequences and their neighbors are fed to the detector model, and their loss under the detector model is measured and used to calculate curvature and to distinguish between generations and human written text.

\noindent\textbf{Curvature (local optimality) test.} The method we use to distinguish between machine-generated and human-written text relies on the local optimality (curvature) of the target sequence, building on the intuition that generations are more likely to be locally optimal than unseen human-written text~\cite{mitchell2023detectgpt,mattern2023}. 
%
%
%To visualize the local neighborhood of the target sequence, we generate perturbations of it and have the target generative model evaluate their loss.
%
%As such, the curvature is then calculated as:

To estimate local optimality, following~\citet{mattern2023,mitchell2023detectgpt}, we generate additional utterances in a local neighborhood around the target by perturbing the target (e.g. re-sampling words at several positions). Then, the measure of local optimality is computed by comparing the likelihood of the target with the likelihood of the local perturbations as follows:

\begin{equation}
   d(x) = \log \ p_{\theta}(x) - \frac{1}{k}\sum_{i=1}^k\log\ p_{\theta}(\Tilde{x_i})
\end{equation}

Here, $x$ is the target sequence, $\theta$ are the parameters of the detector model, and $\Tilde{x_i}$ is the $i$th perturbation of the target utterance $x$ (i.e. the $i$th neighbor) out of the overall $k$ perturbations. The perturbed sequences are generated by  masking parts of $x$ and filling the mask using a perturbation model. The curvature is thresholded to make the machine-generated/human-written text decision.
While technically this measure is an approximate estimate of local optimality, past work has referred to it as 'curvature'. For simplicity, we use this nomenclature going forward.

\noindent\textbf{Perturbation model.}
This model helps generate neighbors by filling in randomly selected spans of the target sequences in the pool and perturbing them. 
We use \textit{T5-3B} for this purpose in our experiments.

\noindent\textbf{Success metric.}
We evaluate the success  of the detector by measuring the area under the ROC curve (AUC), i.e. the false positive vs. true positive rate curve.  The higher the AUC, the more distinguishing power the detection mechanism has.

\noindent\textbf{Evaluation strategy.}
The results we report in the paper fall into two main categories: (1) using a model to detect its own generations, which is the main goal of~\citet{mitchell2023detectgpt} (in this setup, the target and detector models are the same, we call this~\textit{self-detection}); and (2) using a model different from the generator of the text to detect the generations. In this setup, what we are basically doing is acting as if a surrogate model has generated the text. In other words, we want to see how well a model would claim another model's generation as its own. We call this~\textit{cross-detection}.
%
%This second setup represents the black-box case where we not only do not have full access to the target model that generated the text, we also do not know what model it was or what architecture/size it had, so we are trying to find the best~\textbf{universal detector} that would correctly classify it. 

%beyond letting models score their own generated text, which we refer to as \emph{self-detection}, \citet{mitchell2023detectgpt} have also briefly explored the possibility of detecting text generated from other models, which we refer to as \emph{cross-detection}. Due to it being a more realistic setting, we are specifically exploring the cross-detection scenario.

\section{Experimental Setup}
\label{sec:exp}

This section briefly covers the experimental setup. For more details refer to Appendix~\ref{app:exp_setup}.

\noindent\textbf{Models}\label{sec:exp:models}
We use the following model families in our experiments: Facebook's OPT (we use the 125M, 350M, 1.3B, and 6.7B models), EleutherAI's GPT-J, GPTNeo and Pythia~\cite{biderman2023pythia} (we use GPTNeo-125M, GPTNeo-1.3B,  GPTNeo-2.7B, GPTJ-6B and Pythia models ranging from 70M to 2.8B parameters), and OpenAI's GPT models (distilGPT, GPT2-Small, GPT2-Medium, GPT2-Large, GPT2-XL, GPT-3 and ChatGPT). 
%

%We also have experiments where we use partially trained models as detectors. For those experiments, we only use the Pythia models as they are the only ones with available, open-source partially trained checkpoints.
%
%For each Pythia models, there is also a de-duplicated version available, where the model is trained on the de-duplicated version of the data, as opposed to the original dataset.
%
%All the models we use are obtained from HuggingFace~\cite{hf}.

%\subsection{Dataset}

%

\noindent\textbf{Evaluation dataset.}
We follow~\citet{mitchell2023detectgpt}'s methodology for pre-processing and feeding the data to the detector model. 
We use a subsample of the SQuAD~\cite{rajpurkar2016squad} and WritingPrompts~\cite{fan2018hierarchical} datasets, where the original dataset sequences are used as the human-written text in the target sequence pool. We then use the first $20$ tokens of each human-written sequence as a prompt, and feed this to the target model, and have it generate completions for it. We then use this mix of generations and human-written text to create the target pool on which we do the detection.  
In all cases,  our pool consists of $300$ human-written target samples, and $300$ machine-generated samples, so the overall pool size is $600$.

\noindent\textbf{Pre-training datasets for the generator models.}
The {ElutherAI} and {Facebook} models (GPTJ, GPTNeo, Pythia and OPT families)  are all trained on the Pile dataset~\cite{gao2020pile}.
There is limited information and access to the training data of the \textit{OpenAI} models. The GPT-2 family is reportedly trained on the WebText dataset, GPT-3 is trained on a combination of the Common Crawl, WebText2, books and Wikipedia, and there is not any information released about the training data of ChatGPT.

% \subsection{Metrics}
% \paragraph{AUC}

\section{Does cross-detection work?}\label{sec:size}

%As mentioned before, the main goal of our paper is to study ways in which machine-generated text could be distinguished from human-written text, without access to any auxiliary information about the model that generated the text. 
%
%To this end, 
In this section we conduct an extensive set of experiments where we use $23$ models with different sizes and architectures as detectors of text generated by $15$ other models. The results are averaged over the SQuAD and WritingPrompts dataset. We also experiment with  partially trained checkpoints of the detector models, to see how the detection power of the models changes as the training progresses.

\begin{figure*}[t!]
\centering
\vspace{-1ex}
      \includegraphics[width=0.9\linewidth]{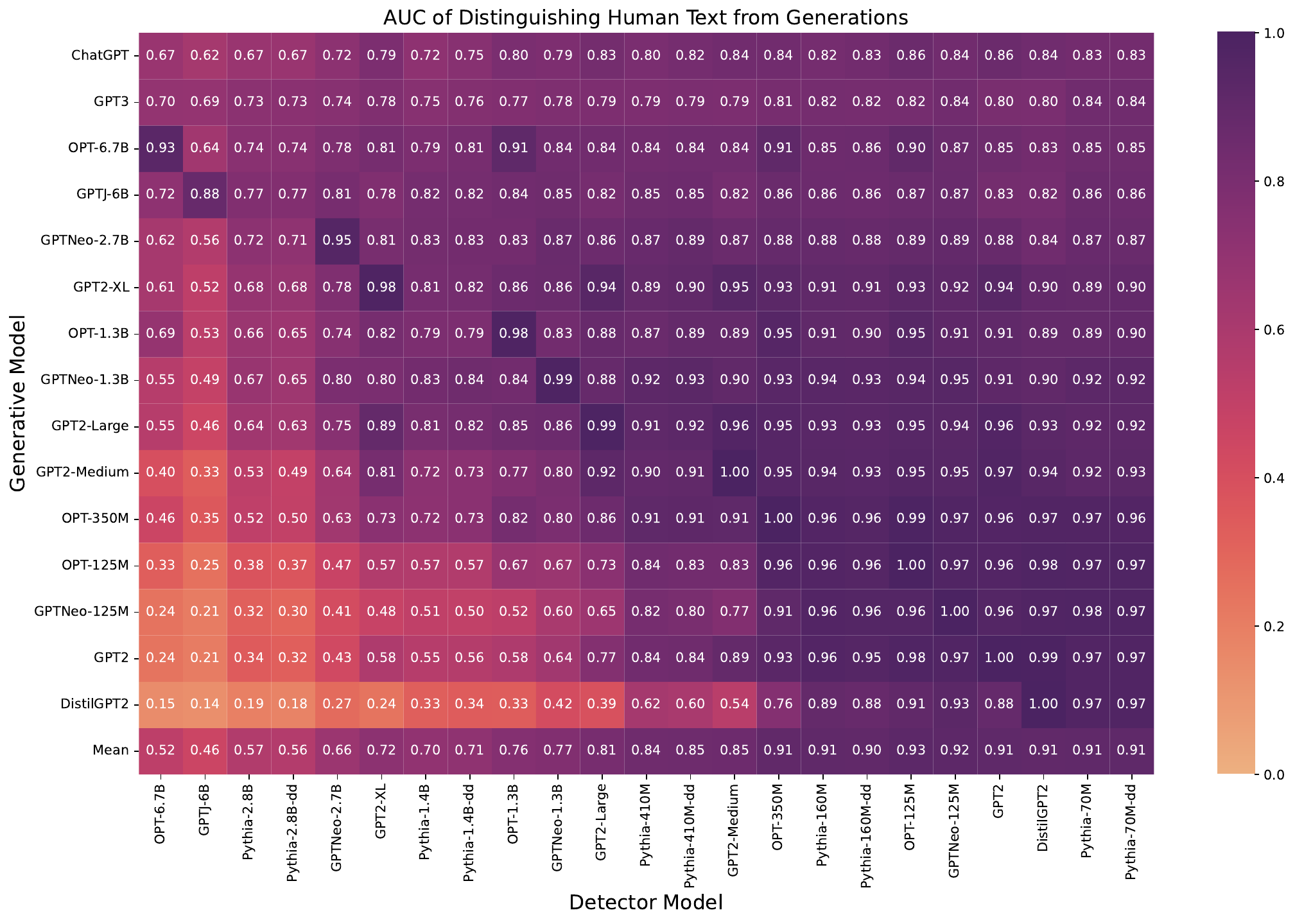}
      \vspace{-1ex}
     \caption{AUC heatmap for cross-detection, where the rows are generator models and columns are the surrogate detector models, both sorted by model size. We can see that smaller models are better detectors and larger models are the worst models in terms of detection power.  }  
       \label{fig:main_heatmap_size}
\end{figure*}

Our main finding is that \textit{cross detection can perform as well as self-detection, or come very close to it}. 
Figures~\ref{fig:main_heatmap_size} and~\ref{fig:summary_heatmap_checkpoint} (full heatmap is Fig.~\ref{fig:main_heatmap_checkpoint} in Appendix) show the AUC of cross-detection for different models.
Figures~\ref{fig:main_heatmap_size_diff} and~\ref{fig:main_heatmap_checkpoint_diff} in Appendix show how close each detector comes, in terms of AUC, to self-detection.
We can see that on average, OPT-125M is the best fully trained universal cross-detector, showing on average $0.07$ lower AUC, compared to self-detection.
If we look at partially trained detector models, however, we see that the Pythia-160M comes as close as $0.05$ AUC points, with its $5k$, $10k$ and $50k$ step trained models (the fully trained model is trained for $143k$ steps). These models  seem to even \textit{outperform} self-detection in some cases, for example when we look at GPTJ-6B generations.
In the rest of this section we further elaborate on these results and draw connections between model size, training, and detection power.

% For instance partially trained Pythia model with 410M parameters can detect GPTJ-6B generations with AUC matching the model itself (See Figure~\ref{fig:main_heatmap_checkpoint_diff}). 

% \begin{table}[]
%     \centering
%     % \footnotesize
%     % \fontsize{7}{7}
% %    \renewcommand{\arraystretch}{0.6}
%     \vspace{-2ex}
%     \begin{adjustbox}{width=0.99\linewidth, center}
%     \input{tabs/summary.tex}
%     \end{adjustbox}
%     \caption{Summary}
%     \vspace{-2ex}
%         \label{tab:summary_size}
% \end{table}

% \begin{figure}[t!]
% \centering
%       \includegraphics[width=0.9\linewidth]{figs/barplot.pdf}
%      \caption{Summary of the cross-detection area under the ROC curve (AUC) results for a  selection of generative (the $4$ models over the X axis) and detector (OPT-125M and OPT-6.7B) models. We can see that the smaller OPT model is a better universal cross-detector. Full results are shown in Figure~\ref{fig:main_heatmap_size}.}  
%        \label{fig:main_barplot_summary}
% \end{figure}
% %%%%%%%%%%%%%

\subsection{Smaller Models Are Better Detectors}\label{sec:smaller}

%In this section we aim to find patterns in the cross-detection power of different models, 
In this section we aim to see if there are any correlations between model size and detection power. 
%we aim to answer the following question: ~\textit{is there any correlation between the size of the model we use to detect machine-generated text, and the detection power?}, for generations from different models.
%
To this end, we use $23$ different models with different parameter counts, ranging from $70M$ to $6.7B$ to detect machine-generation texts from all the models listed in Section~\ref{sec:exp}.
%%%%%%%
%%%%%%%%
%We need to keep in mind that since we are running the curvature test (from Section~\ref{sec:methodology}) using these detector models, we are essentially trying to determine if the detector model could have generated each sequence in the target pool or not.
%

%%%%%%%%%%

\begin{figure*}[]
    \centering
    \begin{subfigure}{0.325\linewidth}
    \centering
     \includegraphics[width=\linewidth]{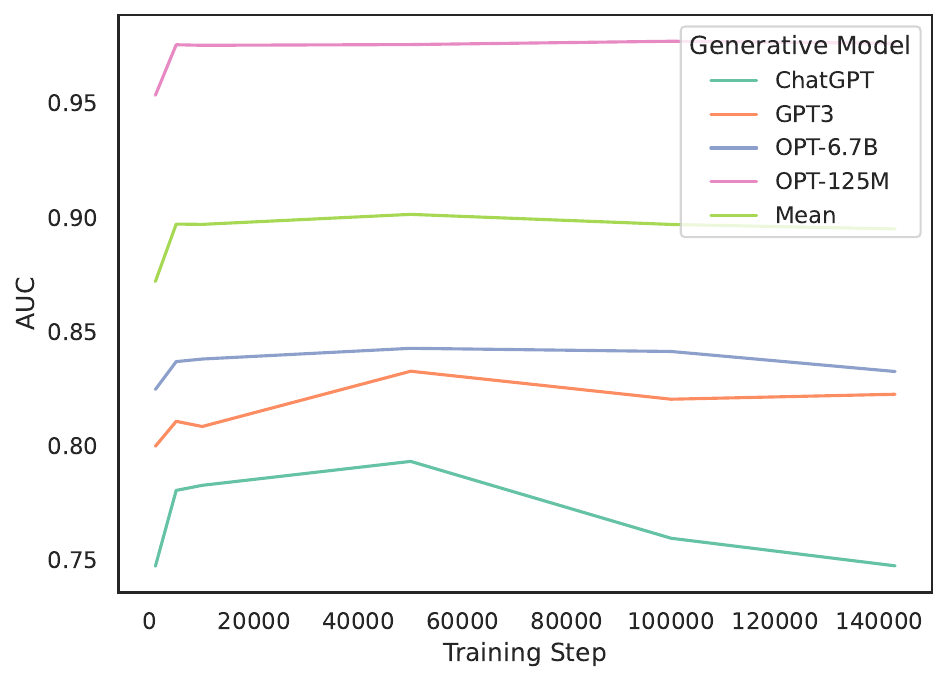}
     \footnotesize
     \caption{Pythia 70M}
     \label{fig:steps_70}
    \end{subfigure}
    \begin{subfigure}{0.325\linewidth}
    \centering
     \includegraphics[width=\linewidth]{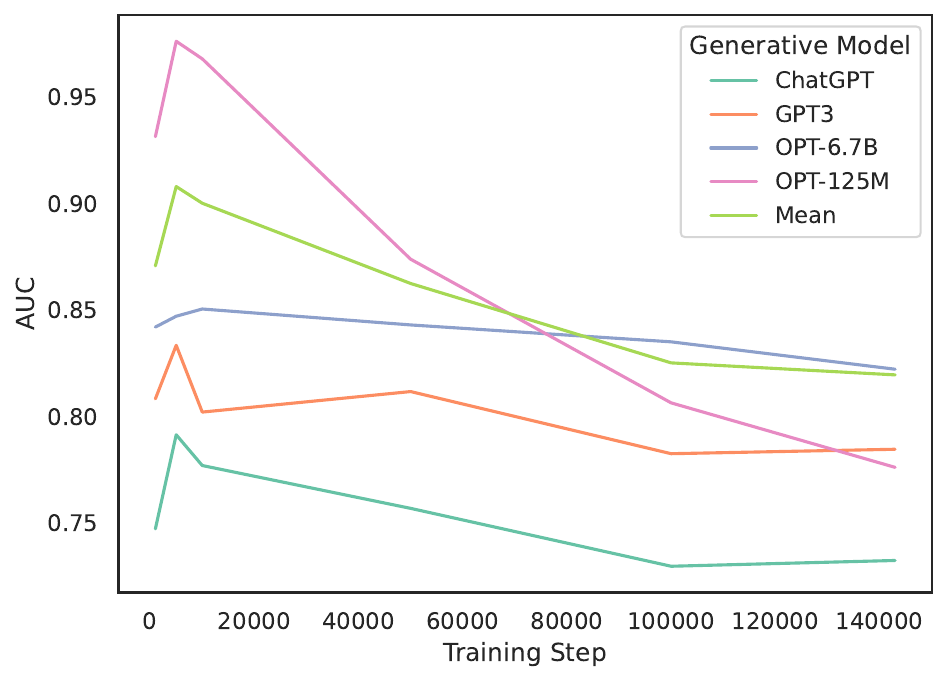}
     \footnotesize
     \caption{Pythia 410M}
     \label{fig:steps_410m}
    \end{subfigure}
    \begin{subfigure}{0.325\linewidth}
    \centering
     \includegraphics[width=\linewidth]{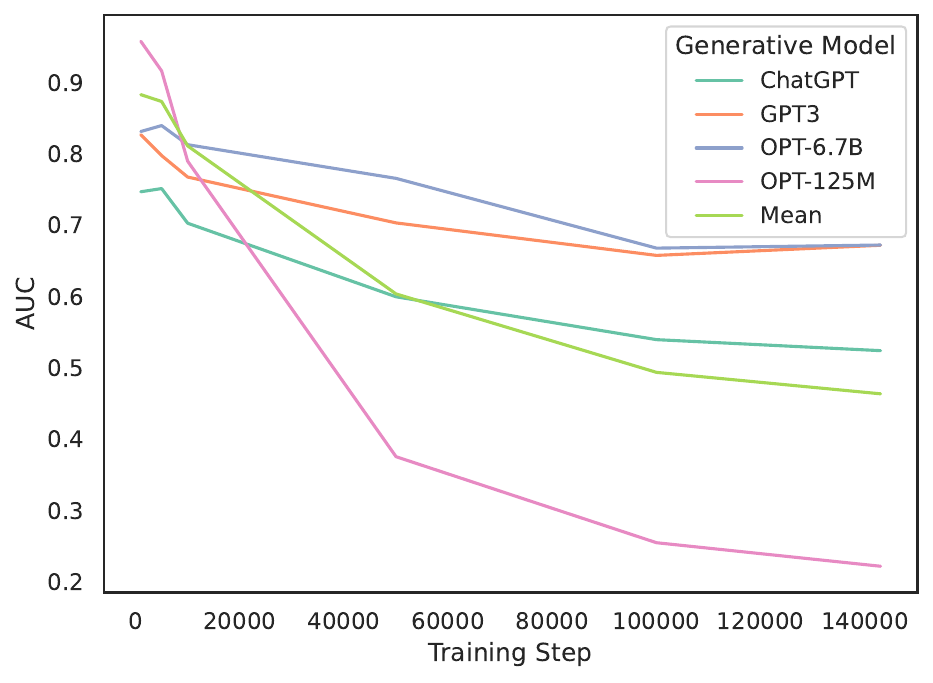}
     \footnotesize
     \caption{Pythia 2.8B}
     \label{fig:steps_2.8B}
    \end{subfigure}
    \caption{Summary of the results for cross-detection power of different detector models trained for different number of steps. Each subfigure shows a different detector model, and the x-axis shows the training step for the checkpoint used as a detector. The results for all $15$ generator models are shown in Figure~\ref{fig:main_heatmap_checkpoint}.}
    \vspace{-1ex}
\label{fig:summary_heatmap_checkpoint}
\end{figure*}

%%%%%%%%%%%%%%%%

%%%%%%%%
\begin{figure*}[]
    \centering
   \begin{subfigure}{0.32\linewidth}
     \centering   
     \includegraphics[width=\linewidth]{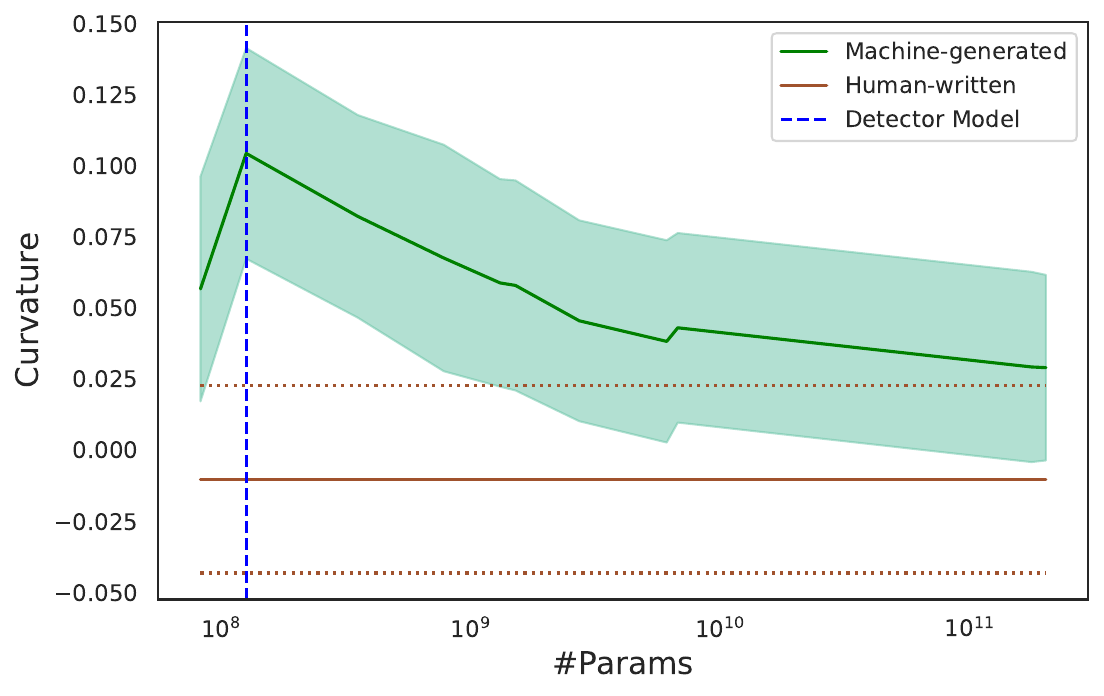}
     \footnotesize
     \caption{Curvature: OPT-125M as Detector}
     \label{fig:curv_good}
    \end{subfigure}
  \begin{subfigure}{0.32\linewidth}
    \centering
     \includegraphics[width=\linewidth]{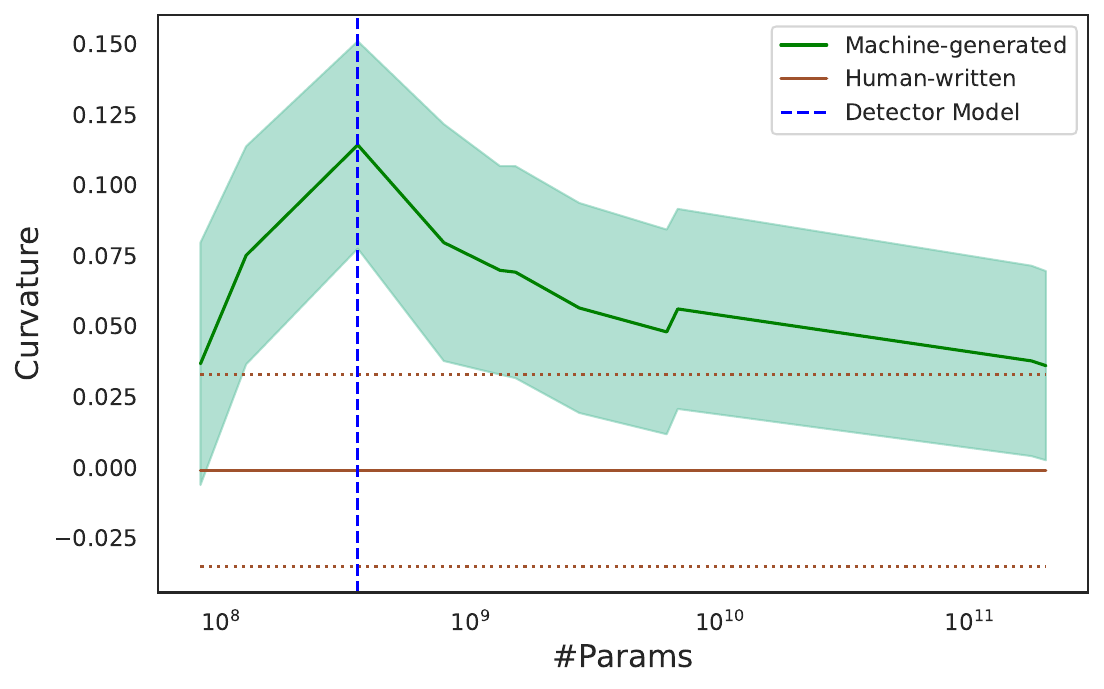}
     \footnotesize
     \caption{Curvature: OPT-350M as Detector}
     \label{fig:curv_med}
    \end{subfigure}
  \begin{subfigure}{0.32\linewidth}
    \centering
     \includegraphics[width=\linewidth]{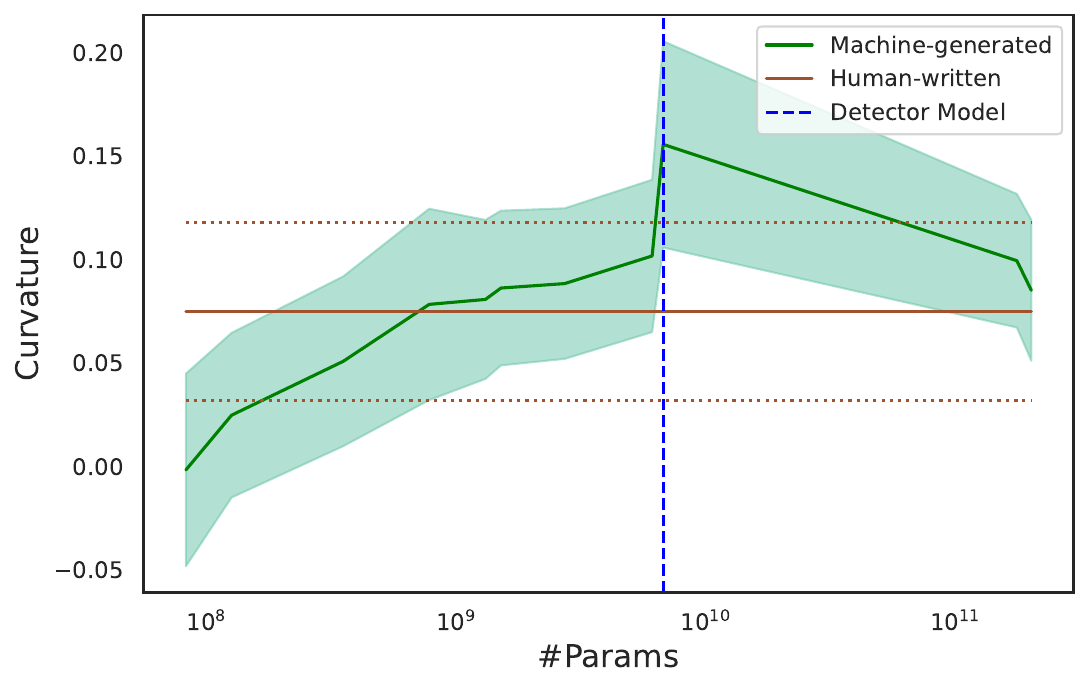}
     \footnotesize
     \caption{Curvature: OPT-6.7B as Detector}
     \label{fig:curv_bad}
    \end{subfigure}

  \begin{subfigure}{0.32\linewidth}
    \centering
     \includegraphics[width=\linewidth]{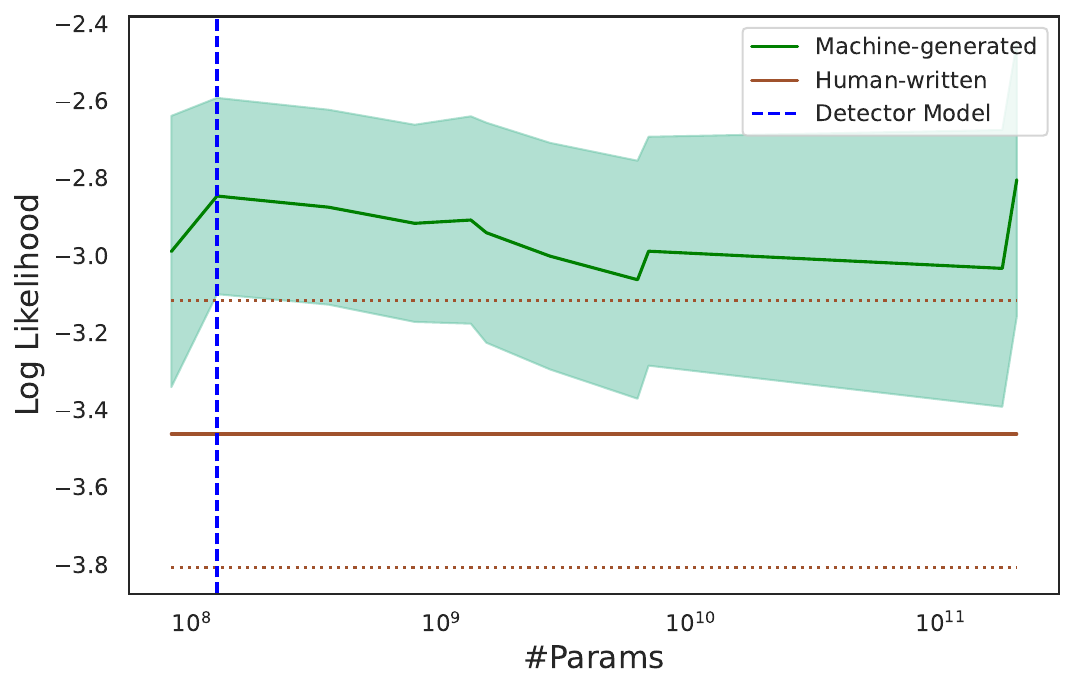}
     \footnotesize
     \caption{Loglikelihood: OPT-125M as Detector}
     \label{fig:ll_good}
    \end{subfigure}
  \begin{subfigure}{0.32\linewidth}
    \centering
     \includegraphics[width=\linewidth]{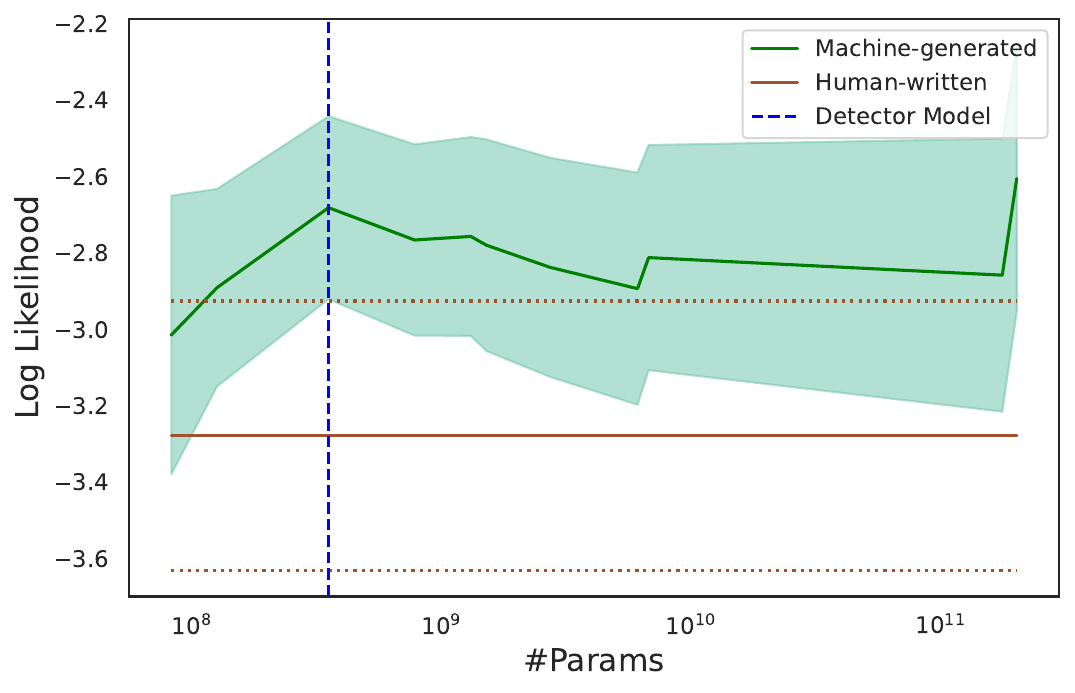}
     \footnotesize
     \caption{Logliklihood: OPT-350M as Detector}
     \label{fig:ll_bad}
    \end{subfigure}
  \begin{subfigure}{0.32\linewidth}
    \centering
     \includegraphics[width=\linewidth]{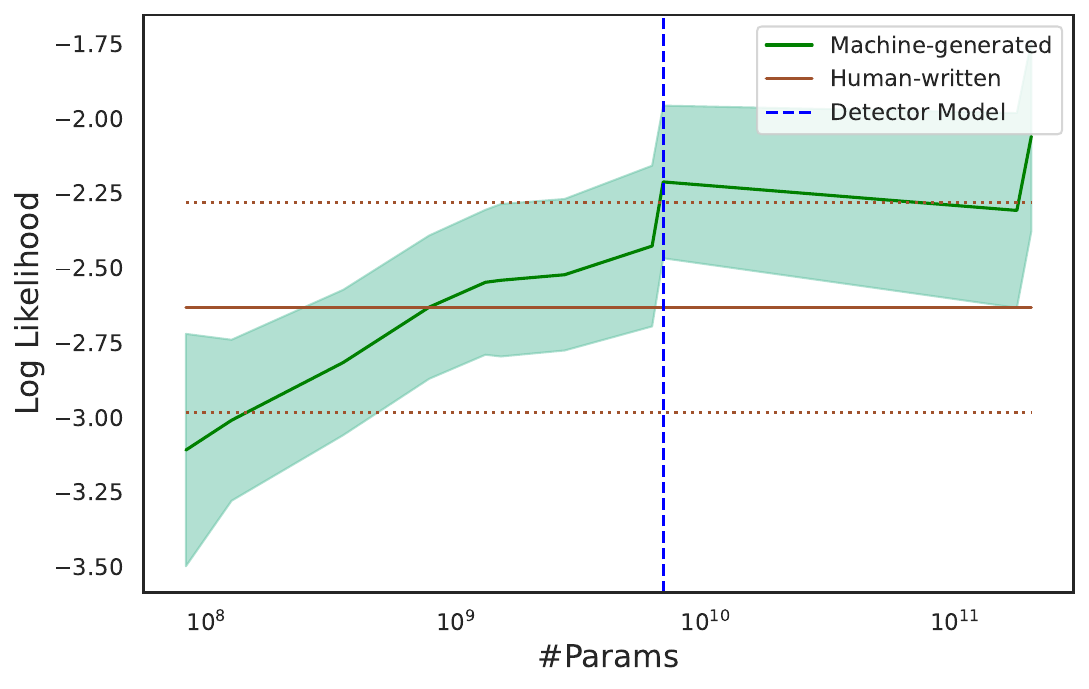}
     \footnotesize
     \caption{Logliklihood: OPT-6.7B as Detector}
     \label{fig:ll_bad}
    \end{subfigure}

\caption{Comparison of curvature and log likelihood values (mean and standard deviation) for the best  universal detector (OPT-125M), a medium sized detector (OPT-350M), and a larger detector from the same family (OPT-6.7B) on generations from models of various sizes (x-axis). The 'Detector Model' line shows values for when the generator and detector are the same model. Detectors tend to show higher curvature on generations than human-written text only for generations from models of the same size or larger. }
    
    \vspace{-2ex}
    \label{fig:curv_auc_bad_good}
\end{figure*}

%%%%%%%%

Figure~\ref{fig:main_heatmap_size} shows the results for this experiment, where the rows are the generator models (sizing up from bottom row to top) and the columns show the detector models (sizing up from right to left). So each cell shows the detection power (AUC) of the given detector model (column), on text generated from the generator model (row). The last row is the mean, which is an overall metric of how good of a detector that model is. 
%
%Figure~\ref{fig:main_heatmap_size_diff} shows how  cross-detection fares against self-detection, and it is missing the ChatGPT and GPT-3 rows as we do not have self-detection results for them (given how we have no access to their loss, or the losses are behind a paywall).

% For both plots, w
We see that the bottom left has the lowest values, showing that \textit{larger models are not good at detecting machine generated text from other models}, and they are particularly bad at it for detecting small model generations.
We can also see that \textbf{smaller models are much better detectors}, as the right side of the graph has much higher AUC values. 
Another observation is the correlations between the \textbf{dataset} and \textbf{model architecture} of the generator and detector models.
%are also important factors when it comes to detecting text generated by other models.
%
As the heatmap shows, models from the same \textit{architecture family} and trained on the same/overlapping \textit{dataset} are better at detecting their own text, compared to models from a different family. 
For instance, for detecting text generated by OPT-6.7B the other models from the OPT family are the best cross-detectors, with AUCs ranging from 0.89-0.87 (OPT-6.7B self-detects with AUC 0.91). The next best cross-detector is the smallest GPTNeo-125M with AUC 0.86. However, the OpenAI GPT2 model of the same size has a lower AUC of 0.84 (and overall the GPT2 family has the lowest cross-detection AUC on OPT), which we hypothesize is due to the larger gap in the training data, as the OPT and GPTNeo/GPTJ models are all trained on the Pile dataset, but GPT2 is trained on the Webtext.
%
%This also explains why overall the OPT/Pythia/GPTJ/GPTNew models are better at cross-detecting OpenAI GPT2 generations, than the other way around, as models are worse at detecting other models' generations, as their 
%
All in all, the difference due to the dataset/architecture differences is small as most of the dataset for all these models is comprised of web-crawled data, showing that cross-detection can be effective, regardless of how much information we have about the target model, and how accessible similar models are.

%We have also provided an overall summary of the heatmaps in Figure~\ref{fig:main_barplot_summary}, where we have presented the numbers from the best overall detector with mean AUC of 0.92 (OPT-125M) and the biggest model of the same family, OPT-6.7B with average AUC of 0.46.
%
One noteworthy observation is that OPT-125M can detect generations from models like GPT3 and ChatGPT with relatively high AUC (0.81). However, if the intuitive approach of taking another large, ``similar'' model were to be taken and we were to use OPT-6.7B, we would get AUC of 0.67 and 0.58 for these models, respectively, which are both close to random (0.5). 
%
% We hypothesize that the reason behind large models being poor detectors of text generated by other models (especially smaller ones), is that larger models have a more refined taste, therefore they  don't attribute text generated by other models as their own generations.  Smaller models, however, attribute any machine-generated text as their own, since they have a less specific taste and are looser fitting models. 
Thus, intuitively, it seems that larger models have more refined taste: they only show higher local optimality (relative to human-written text) on generations from large models. Conversely, smaller models are more forgiving: they show higher local optimality on generations from similarly small models and larger, making them better universal detectors via local optimality comparison.
We discuss this further in Section~\ref{sec:curvature}.
% \paragraph{Large models are bad cross detectors.}

% \paragraph{Small models are good cross detectors.}

\subsection{Partially Trained Models are Better Detectors}

% \begin{table}[]
%     \centering
%     % \footnotesize
%     % \fontsize{7}{7}
% %    \renewcommand{\arraystretch}{0.6}
%     \vspace{-2ex}
%     \begin{adjustbox}{width=0.99\linewidth, center}
%     \input{tabs/summary_chkpt}
%     \end{adjustbox}
%     \caption{Summary table partially trained}
%     \vspace{-2ex}
%         \label{tab:summary_checkpoint}
% \end{table}

%%%%%%%%%%%%%

%%%%%%%%%%

%Our approach in this section is very similar to the previous one, except here we aim to find correlations between how far along in the training process a model is, and its cross-detection power.
%for generations from different models.
%
%To this end, 
We take different training checkpoints of the Pythia models~\cite{biderman2023pythia} at different steps (steps $1k$, $5k$, $10k$, $50k$, $100k$ and $143k$) with different sizes (2.8B, 410M, and 70M), and use them as detectors of generations from the 4 target models.
Figure~\ref{fig:summary_heatmap_checkpoint} shows the results for this experiment (Figures~\ref{fig:main_heatmap_checkpoint} and~\ref{fig:main_heatmap_checkpoint_diff} show entire heatmaps of this experiment).
%, with the rows showing the generative models, sizing up from the bottom to top and the columns shows the detector models at different training checkpoints, where from left to right the models size down and are also trained for fewer iterations. 
%
For each model we can see that ~\textbf{the final checkpoint is consistently the worst one in terms of machine-generated text detection}, and it is one of the middle checkpoints that has the best performance. 
Our hypothesis for this is similar to that of Section~\ref{sec:size}, where we believe that partially trained models have not yet fit to the training data tightly (and have a smoother surface), so they over claim other models' generations as their own, whereas the longer a model is trained, the sequences it ranks higher as its own narrow down.

    \begin{figure}
    \centering
\includegraphics[width=0.95\linewidth]{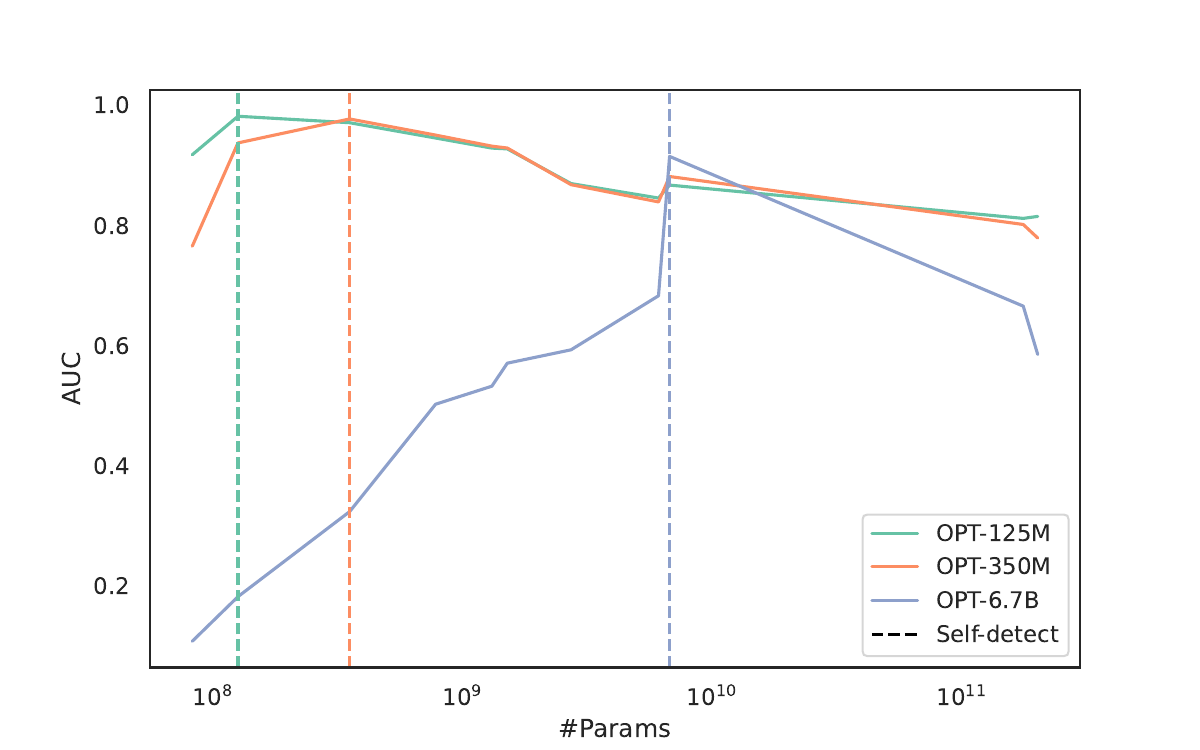}
     \footnotesize
     \caption{AUC of the three cross-detectors from Figure~\ref{fig:curv_auc_bad_good}}
     \vspace{-1ex}\label{fig:auc_bad_good}
    \end{figure}

\section{How are smaller models better detectors? }\label{sec:curvature}
%\subsection{Relationship between Curvature and Detector Size}\label{sec:curvature}

To help shed light on why smaller models are better detectors and larger models are not good at detecting machine generated text, we plot a breakdown of the curvature metric (Section~\ref{sec:methodology}) and log-likelihood values for the best universal detector (OPT-125M), a medium sized detector of the same family (OPT-350M) and a larger one from the same family (OPT-6.7B), shown in Figure~\ref{fig:curv_auc_bad_good}.
The y-axis is the curvature/log likelihood of the target generations (from the $15$ models from Section~\ref{sec:exp:models}) under the detector models (OPT-125M, 350M or 6.7B).  The x-axis is the number of parameters of the generator model (we do not know how many parameters ChatGPT has, so we plotted it as the rightmost point in the plots). 
Figure~\ref{fig:auc_bad_good} plots the AUCs for detection under the three models, for the $15$ generator models. 

%, which are the OPT-125M and GPT-J 6B models, respectively. 
%
We can see that for the smaller detector model (Figures~\ref{fig:curv_good} and~\ref{fig:ll_good}), the mean curvature and log-likelihood values for the generated text are consistently higher than the curvature for the human-written text. However, for the larger model (Figure~\ref{fig:curv_bad} and~\ref{fig:ll_bad}), the curvature and log-likelihood values for the machine-generated text is in most cases smaller than or around the same value as the human written text.
The curvature and log-likelihood values for human written text for both graphs are stable since the text is the same and doesn't depend on the target model.

We can also see that overall the curvature and likelihood values  for the larger model are higher, especially for the original text, than those of the smaller model, and the values for text generated by the other models have lower curvature and likelihood value. This shows that the larger model places higher likelihood on the human written text and fits it better.
The smaller model, however, assigns lower curvature  and likelihood to the human-written text compared to generations by a large gap, and the assigned values are overall lower than those of the large model. 
Broadly we observe that  \textbf{all models respond similarly to machine generated text from other models, so long as the other model is same size or bigger.} In other words, they place high likelihood on text from larger models. However, for models smaller than themselves, they place lower likelihood and curvature. As such, smaller models are \textbf{better universal detectors}, as the size of the set of sequences they assign higher likelihood and curvature to is bigger than it is for large models, and this higher curvature is much higher than the curvature assigned to the human written text.
The spikes in all the sub-figures of Figure~\ref{fig:curv_auc_bad_good} graphs are for the detector model detecting its own text.

\section{Does neighborhood choice matter? }

Our estimation of ``curvature'' hinges upon generating numerous perturbations (neighbors) and comparing their loss with that of a target point. Therefore, if these perturbed neighbors are not sufficiently nearby and lie in a different basin of the likelihood surface, our measure of curvature is not  accurate (the closer the perturbed points are, the more accurate estimation of curvature we achieve).
The perturbation method directly impacts the size and shape of the neighborhood we create.  Therefore, we compare different perturbation schemes in order to see how sensitive detectors of different sizes are to neighborhood choice.
%around a target point, and use to determine the shape of the loss function around it and test its local optimality.
%
%~\textbf{If the generated perturbations are too far from a target point, they will have lower likelihood and create inaccurately high curvature estimates}.

%As mentioned in Section~\ref{sec:size}, one hypothesis we have for why small models are better machine-generated text detectors is that they have flatter, looser fitting loss functions whereas larger models have higher curvatures, are sharper and more compressed.
%
%As such, for better analysis of the shape of a function around a target  point on a larger model, one needs to generate perturbations closer to that point to magnify the local neighborhood where we test optimality, since we hypothesize that the function is more spiked and changes fast, as opposed to a smaller model that is smoother.
%
%To further explore this hypothesis,

We investigate two different methods for changing the distance of the generated perturbations: (1) we change the mask filling model size, by experimenting with \textit{T5-Small}, \textit{T5-Large} and \textit{T5-3B}~\cite{hf,raffel2020exploring} to test the intuition that larger mask-filling models, generate semantically closer neighbors than a smaller model, we present the extended results for this in Appendix~\ref{app:mask-fill}. A similar analysis is also conducted in~\cite{mitchell2023detectgpt}, we however, do a more extensive analysis on numerous models of different sizes and probe the curvature values. (2) We change the percentage of the tokens that get masked and replaced by the mask-filling model, as the more tokens we mask and replace, the farther the generated perturbations would be. (3) Finally, we look into how many tokens we actually need in the generated/human-written sequences to create a neighborhood and be able to accurately distinguish the texts.
%(3) we change the mask span lenght that is being replaced, which determines how many tokens we re-sample at a time to generate neighbors. 

%%%%%%%%%%%%%
%%% Mask coverage pct

\begin{figure*}[]
    \centering
    \begin{subfigure}{0.32\linewidth}
    \centering
     \includegraphics[width=\linewidth]{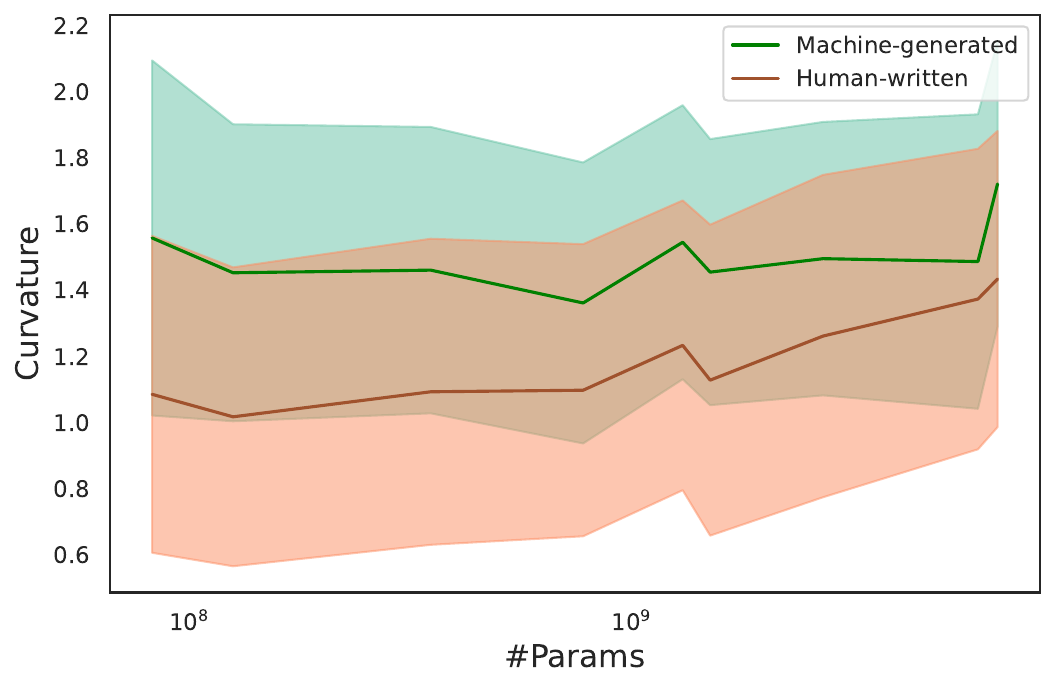}
     \footnotesize
     \caption{$90\%$ Masking}
     \label{fig:msk_9}
    \end{subfigure}
    \begin{subfigure}{0.32\linewidth}
    \centering
     \includegraphics[width=\linewidth]{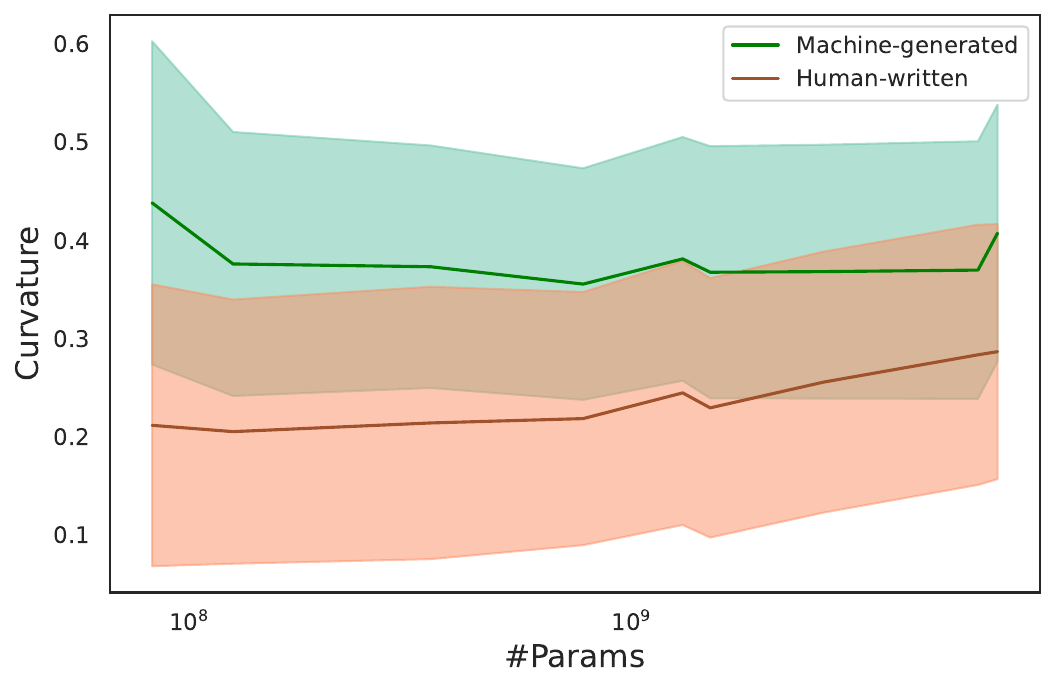}
     \footnotesize
     \caption{$50\%$ Masking}
     \label{fig:msk_5}
    \end{subfigure}
    \begin{subfigure}{0.32\linewidth}
    \centering
     \includegraphics[width=\linewidth]{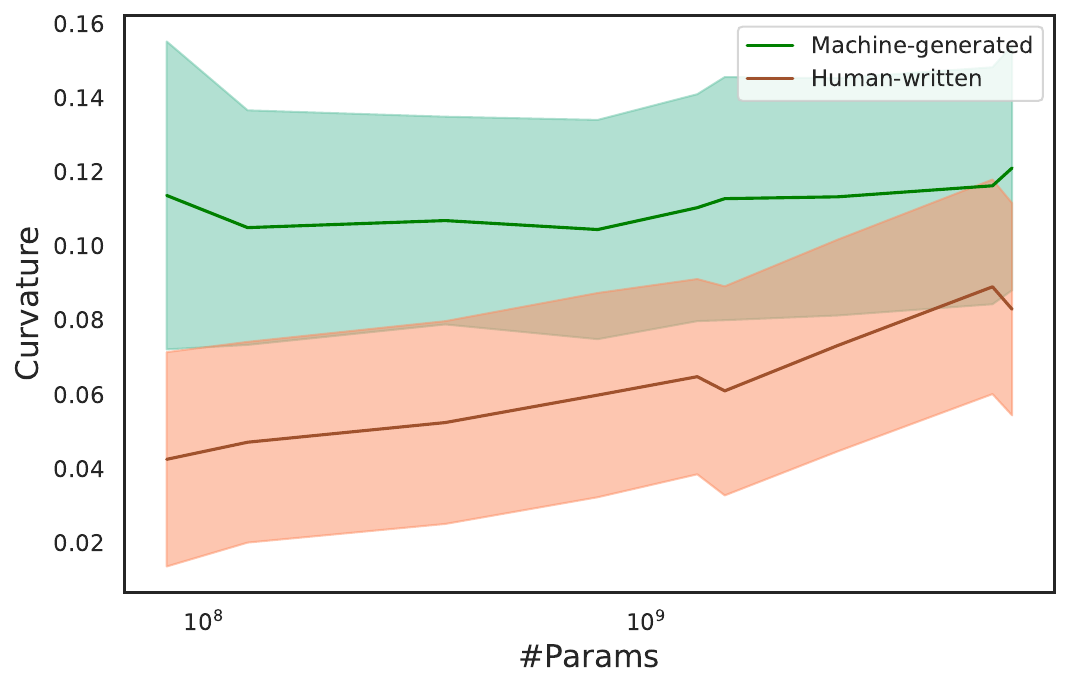}
     \footnotesize
     \caption{$15\%$ Masking}
     \label{fig:msk_15}
    \end{subfigure}

    \begin{subfigure}{0.32\linewidth}
    \centering
     \includegraphics[width=\linewidth]{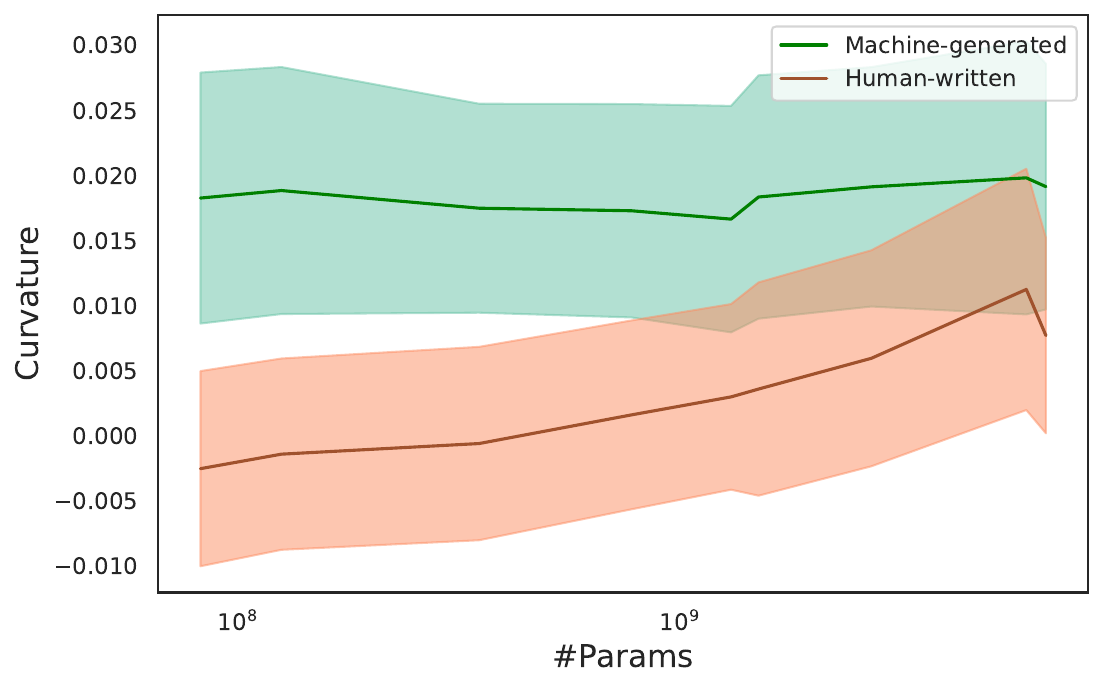}
     \footnotesize
     \caption{$2\%$ Masking}
     \label{fig:msk_2}
    \end{subfigure}
        \begin{subfigure}{0.32\linewidth}
        \centering
     \includegraphics[width=\linewidth]{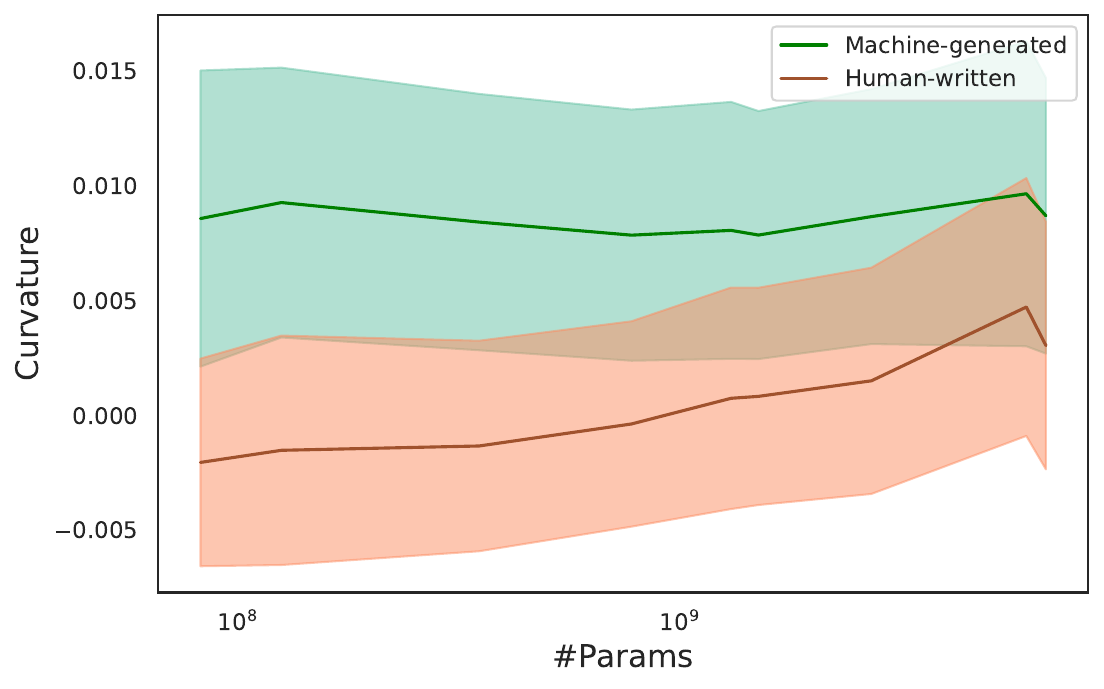}
     \footnotesize
     \caption{$1\%$ Masking}
     \label{fig:msk_1}
    \end{subfigure}
 \begin{subfigure}{0.32\linewidth}
     \includegraphics[width=\linewidth]{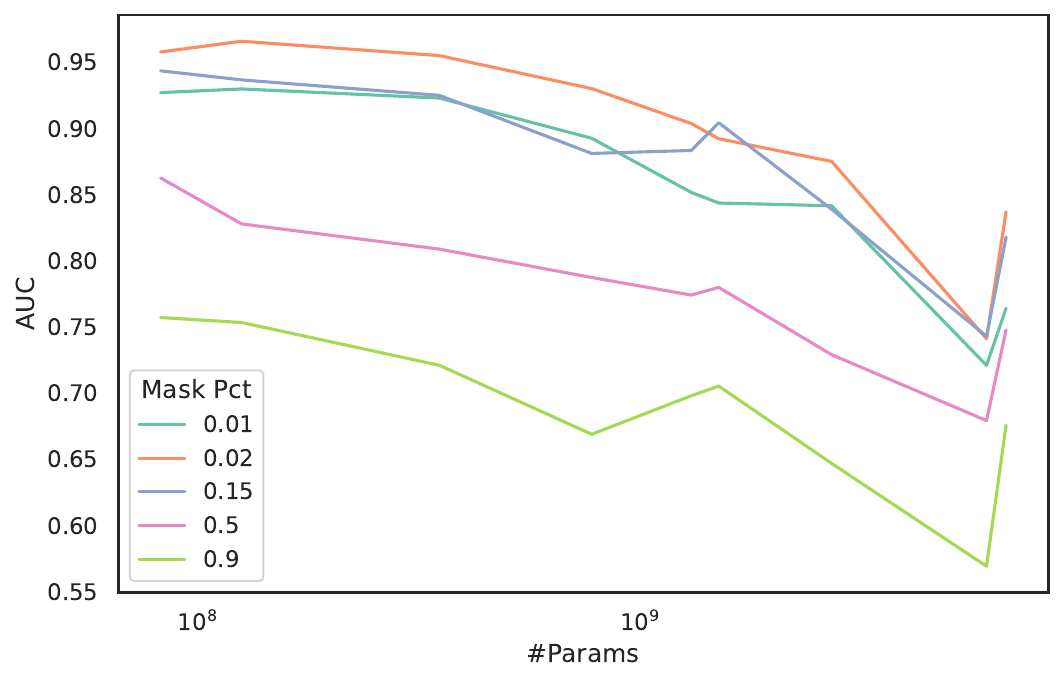}
     \footnotesize
     \caption{AUCs for different masking pctgs.}
     \label{fig:msk_auc}
    \end{subfigure}
    
\caption{The effect of changing the masking percentage on curvature values and self-detection power of different models with different sizes (AUC).}
    \label{fig:msk_pct}
    \vspace{-2ex}
\end{figure*}

%%%%%%%%%%%%%%%%
\subsection{Masking Percentage}
Figure~\ref{fig:msk_pct} shows the results for the experiment where we change the percentage of tokens that are masked, to produce the neighbors. In all previous experiments, we used $15\%$ masking with mask span length of $2$ tokens following the experimental setup in~\citet{mitchell2023detectgpt}.
In this section, however, we change the percentage of the masked tokens (and we set the masking to be contiguous) to see how it affects the curvature mean and standard deviation values, and the AUCs. 
We can see that as the masking percentage decreases (from $90\%$ to $2\%$), the AUCs and the self-detection power of models increase rather consistently. 
When we go to $1\%$, however, we see the AUC drop. If we look at Figure~\ref{fig:msk_1} which depicts the curvature measures for the $1\%$ masking, we see that the curvatures overlap between machine-generated and human-written text, which we hypothesize is because our implementation does not enforce that re-sampled words must differ from the words they are replacing. Thus, for the smallest masking percentage, it is possible that some perturbations are identical to the target, which may explain reduced detection accuracy in this setting\footnote{Its noteworthy that the slight discrepancy between the results for $15\%$ masking in this section and the previous section is that there, the mask span length was $2$ so the masked portion of the sequence is not contiguous. In this experiment, however, we  use contiguous masking. 
}.
%the perturbations are all too close (or even identical) to the original sequences, and as such do not define the neighborhood well.
%

\subsection{How many tokens do we need for detection?}
    \begin{figure}
    \centering
\includegraphics[width=0.95\linewidth]{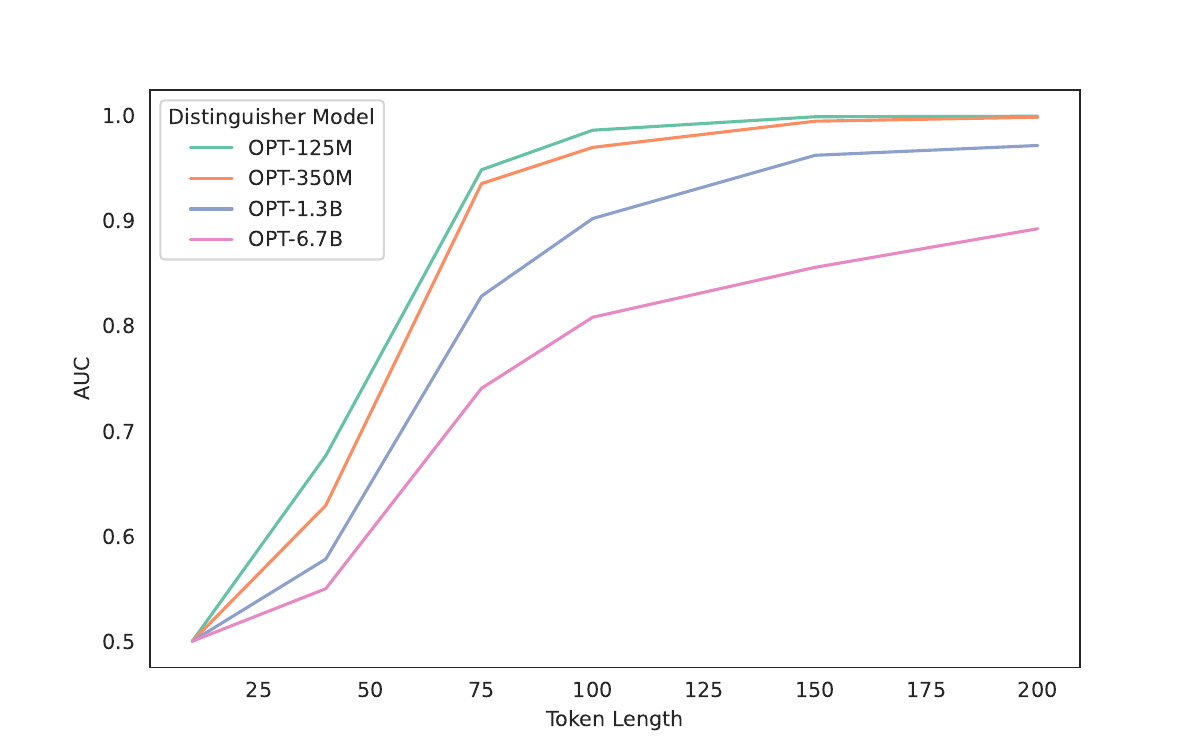}
     \footnotesize
    \caption{Detectability as a function of candidate utterance length. As expected, longer utterances are more cross-detectable -- though it's worth noting that utterances as short as 60 tokens long are still cross-detectable with relatively high accuracy.}
     \label{fig:seqlen}
    \end{figure}

Figure~\ref{fig:seqlen} shows how the length of the target sequence  affects the sequence's detectablity (AUC of detection), and how many tokens we need to be able to do precise detection. We compare sequences of different lengths, ranging from 10 tokens to  200, for four different models with four different parameter counts, on the SQuAD dataset. In this setup we target self-detection. We can see that the longer the sequence, the easier it is to distinguish if it is human-written or machine-generated, and 75-100 tokens seems like the point where we hit diminishing returns.
We can also see that across different sequence lengths, as models get smaller, the detection power increases, as seen throughout the rest of the paper.

%%%%%%%%%%%%%%%%%%%
%\subsection{Span Length}

\section{Related Work}

The problem of machine-generated text detection has already been studied for multiple years using a variety of different approaches \citep{ippolito-etal-2020-automatic, jawahar-etal-2020-automatic, uchendu-etal-2020-authorship, uchendu-etal-2021-turingbench-benchmark}: Both \citet{gehrmann-etal-2019-gltr} and \citet{dugan2022real} have found that humans generally struggle to distinguish between human- and machine-generated text, thereby motivating the development of automatic solutions. Among those, some methods aim to detect machine-generated text by training a classifier in a supervised manner \citep{bakhtin2019real, uchendu-etal-2020-authorship}, while others perform detection in a zero-shot manner \citep{solaiman2019release, ippolito-etal-2020-automatic}. There is also a line of work that relies on bot detection through question answering~\cite{wang2023bot,Chew2003BaffleTextAH}, which is outside the scope of this paper.

 Most recently, \citet{mitchell2023detectgpt} introduced the zero-shot method DetectGPT, which is based on the hypothesis that texts generated from a LLM lie on local maxima, and therefore negative curvature, of the model's probability distribution. 
 %Thus, minor rewrites of machine-generated texts, which are in practice obtained through word replacements suggested by a separate model such as T5 \citep{raffel2020exploring}, are consistently assigned lower probabilities than the original text, whereas rewrites of human-written texts can have both higher or lower probabilities assigned to them.
%
%In many cases, access to model probabilities might not be provided by third party LLM APIs, and furthermore, as a multitude of high quality models are available to the public, it is realistically not always possible to know which model should be used to score candidate texts and their rewrites. Thus, we distinguish between the scenario of self-detection, where the target model used to generate text is at the same time the detector model scoring texts and their rewrites, and the scenario of cross-detection , in which the target model and the detector model are different models. For reasons described above, it would be desirable to be able to reliably detect machine-generated text in the cross-detection scenario. \citet{mitchell2023detectgpt} have already explored this and found that some models work better as detectors while others don't. In this work, we aim to answer thoroughly investigate what makes a model a good detector model.
%
%Beyond the approaches discussed in this paper, 
Other strategies have been proposed to enable the detection of machine-generated text in the wild. Particularly through efforts on the side of the LLM provider, more powerful detection methods can be devised. 
One such method is watermarking, which injects algorithmically detectable patterns into the released text while ideally preserving the quality and diversity of language model outputs. Watermarks for natural language have already been proposed by \citet{10.1007/3-540-45496-9_14} and have since been adapted for outputs of neural language models \citep{fang-etal-2017-generating, ziegler-etal-2019-neural}. Notable recent attempts for transformer based language models include work by \citet{abdelnabi21oakland}, who propose an adversarial watermarking transformer (AWT). While this watermarking method is dependent on the model architecture, \citet{kirchenbauer2023watermark} propose a watermark that can be applied to texts generated by any common autoregressive language model.
As a strategy more reliable than watermarking, \citet{krishna2023paraphrasing} suggest a retrieval-based approach: By storing all model outputs in a database, LLM providers can verify whether a given text was previously generated by their language model. In practice, this would however require storage of large amounts of data and highly efficient retrieval techniques in order to provide fast responses as the number of generated texts grows.

\paragraph{Relationship to Membership Inference Attacks (MIA)}Prior work~\cite{mattern2023} demonstrated that the same optimality test  can be used to distinguish between training set members and non-training members, i.e. as a membership inference attack.
As our experiments showed, when models size up the detection power (i.e. distinguishablity between machine-generated and human-written text) decreases. %We can see that the detection power for the Pile is consistency lower than WritngPrompts, which we attribute to the fact that the Pile is part of the training set so it is harder for all models to distinguish between generations based on it and the actual data.
For MIA, however, prior work demonstrate inverse scaling, as in larger models demonstrate higher distinguishing power~\cite{mireshghallah2022quantifying,mattern2023}.
We attribute this to the higher memorization capablities of these models, as shown by~\cite{tirumala2022memorization}, making it easier for them to recognize their training data. 

\section{Conclusion}

With the increasing prevalence of LLMs and their integration into various different services, it becomes crucial to differentiate between text written by humans and text generated by machines so as to avoid fake news and impersonations.
As such, we set out to explore the possibilities of using existing models to detect generations from unknown sources, and distinguish them from human written text. 
We find that when using zero-shot detection methods that rely on local optimality, smaller models are overall better at detecting generations, and larger models are  poor detectors. 
Our results offer hope of robust general purpose protection against LLMs used with nefarious intentions. However, as LLMs continue to change and detection evasion methods become more prevalent, so must methods for detection and validation studies.

\section*{Limitations}

%We acknowledge that running inference on larger models with bil

%We want to clarify that the results provided in Section~\ref{sec:mia} should be taken with a grain of salt, given the possibility of dataset contamination, as we cannot be completely sure that the WritingPrompts dataset was not entirely used in the training of the OPT models.
%
%We also acknowledge that our definition of curvature is only based on heuristic, 
Although we see high AUCs for black-box detection of machine generated text in our experiments, this does not necessarily mean that these detection methods are not avoidable, and that they can be applied to all models and achieve high performance. Further experiments are needed to evaluate the generalization of our findings to other architectures and setups.

\section*{Acknowledgments}

 This research is supported in part by DARPA SemaFor Program No. HR00112020054. We thank Yejin Choi, Kaj Bostrom, and members of UW NLP and UCSD BergLab for insightful discussions.

\bibliography{anthology,main}
\bibliographystyle{acl_natbib}

\appendix

%%%%%%%%
\clearpage 
\appendix  
\section{Ablating Mask Filling Models}\label{app:mask-fill}

%%%%%%
\begin{figure*}[t!]
\centering
\includegraphics[width=0.95\linewidth]{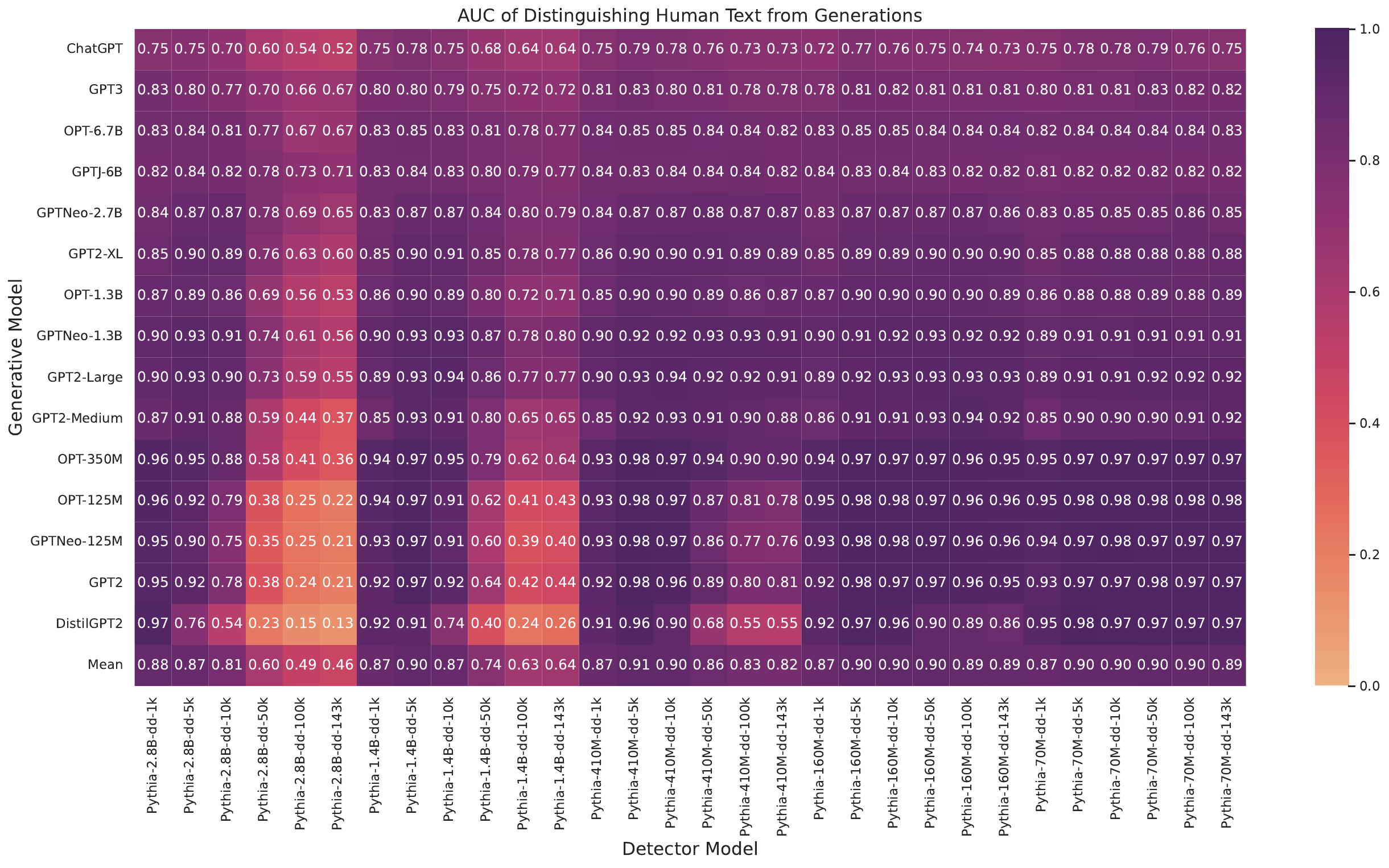}
     \caption{AUC heatmap for cross-detection, where the rows are generative models and columns are the surrogate detector models from the Pythia family, at different training step checkpoints ($1k$, $5k$, $10k$, $50k$, $100k$ and $143k$), both sorted by model size. We can see that partially trained models are better detectors. }  
\label{fig:main_heatmap_checkpoint}
\end{figure*}
%%%%%%%
\begin{figure*}[t!]
\centering
      \includegraphics[width=0.95\linewidth]{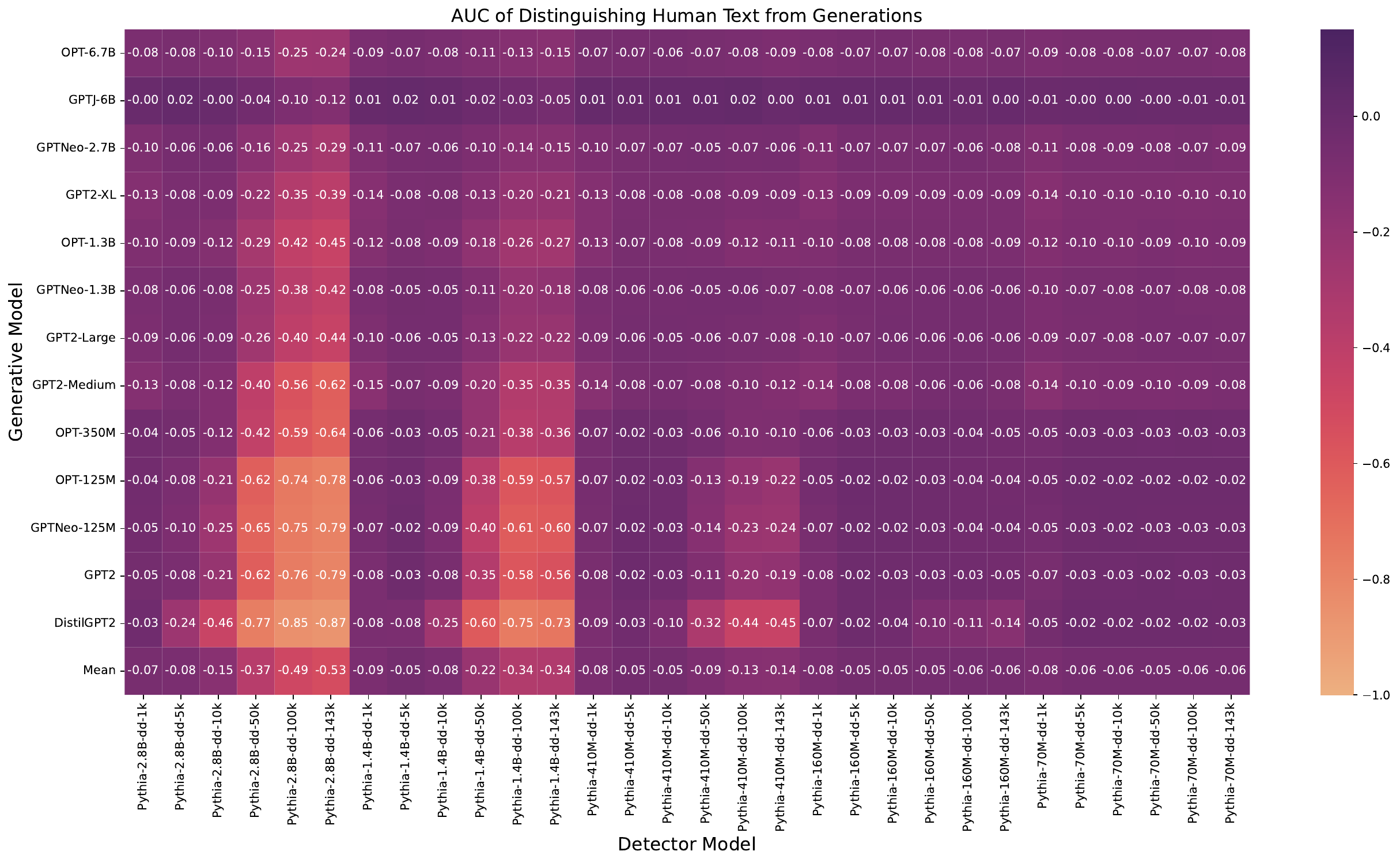}
     \caption{AUC difference between self-detection and cross-detection heatmap (to better see how close cross-detection comes to self detection), here the rows are generative models and columns are the surrogate detector models from the Pythia family, at different training step checkpoints ($1k$, $5k$, $10k$, $50k$, $100k$ and $143k$), both sorted by model size. This plot is basically Figure~\ref{fig:main_heatmap_checkpoint}, where each cell in a row is subtracted by the self-detection AUC for that row.} 
\label{fig:main_heatmap_checkpoint_diff}

\end{figure*}
%%%%%%%

\begin{figure*}[t!]
\centering
      \includegraphics[width=0.95\linewidth]{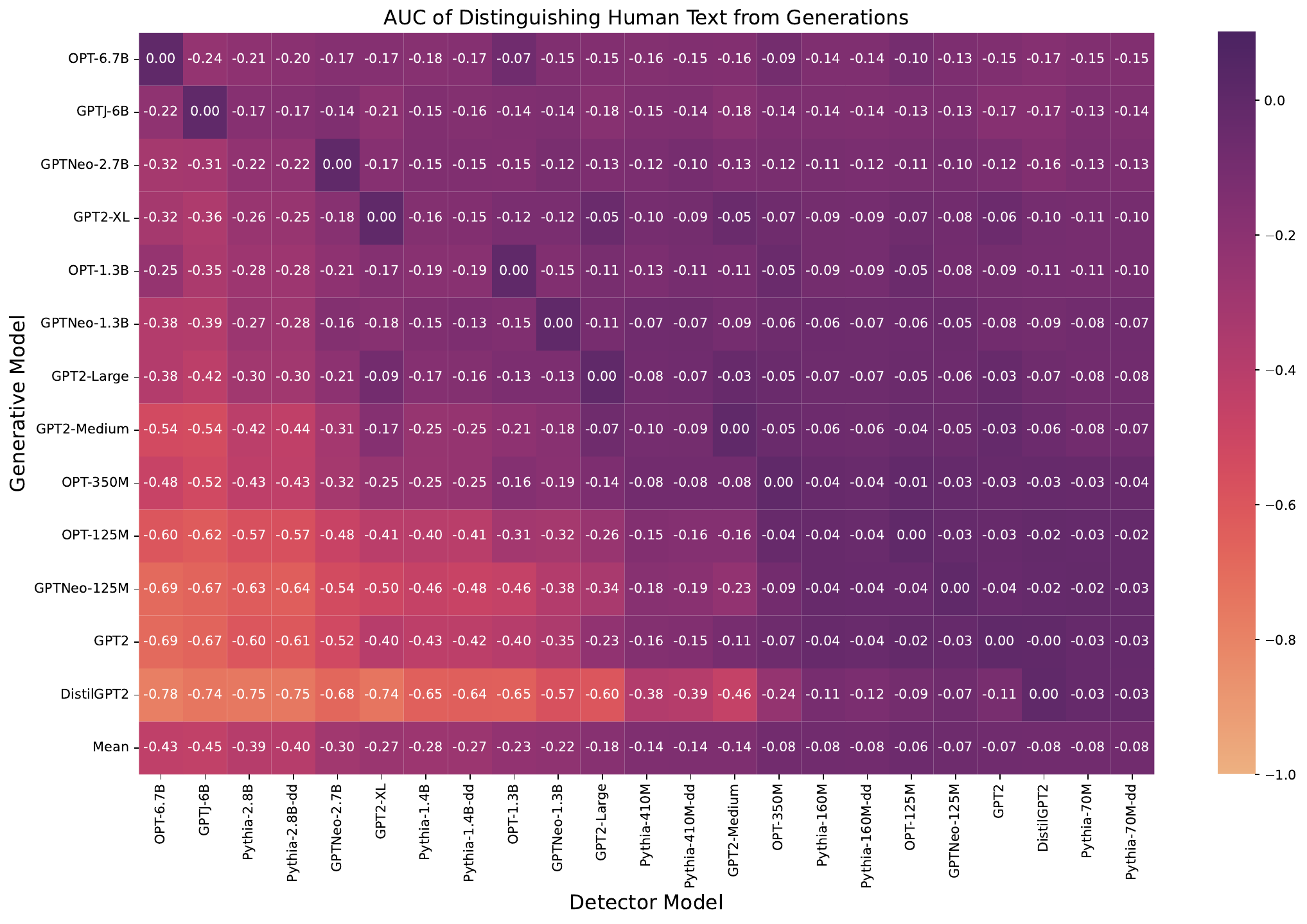}
     \caption{AUC difference between self-detection and cross-detection heatmap (to better see how close cross-detection comes to self detection), where the rows are generative models and columns are the surrogate detector models, both sorted by model size. This plot is basically Figure~\ref{fig:main_heatmap_size}, where each cell in a row is subtracted by the self-detection AUC for that row.} \label{fig:main_heatmap_size_diff}
\end{figure*}

%%%masking model
%%% masking model

\begin{figure}[]
    \centering
    \begin{subfigure}{\linewidth}
        \centering
     \includegraphics[width=0.8\linewidth]{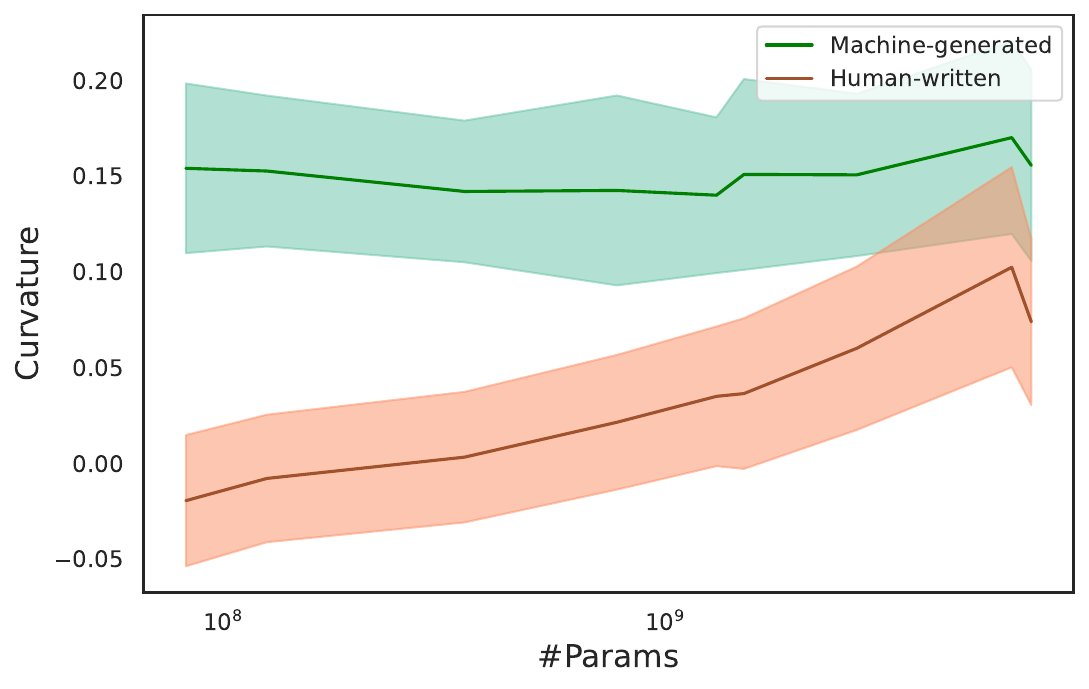}
     \footnotesize
     \caption{T5-3B}
     \label{fig:t5-3b-dist}
    \end{subfigure}

    \begin{subfigure}{\linewidth}
        \centering
     \includegraphics[width=0.8\linewidth]{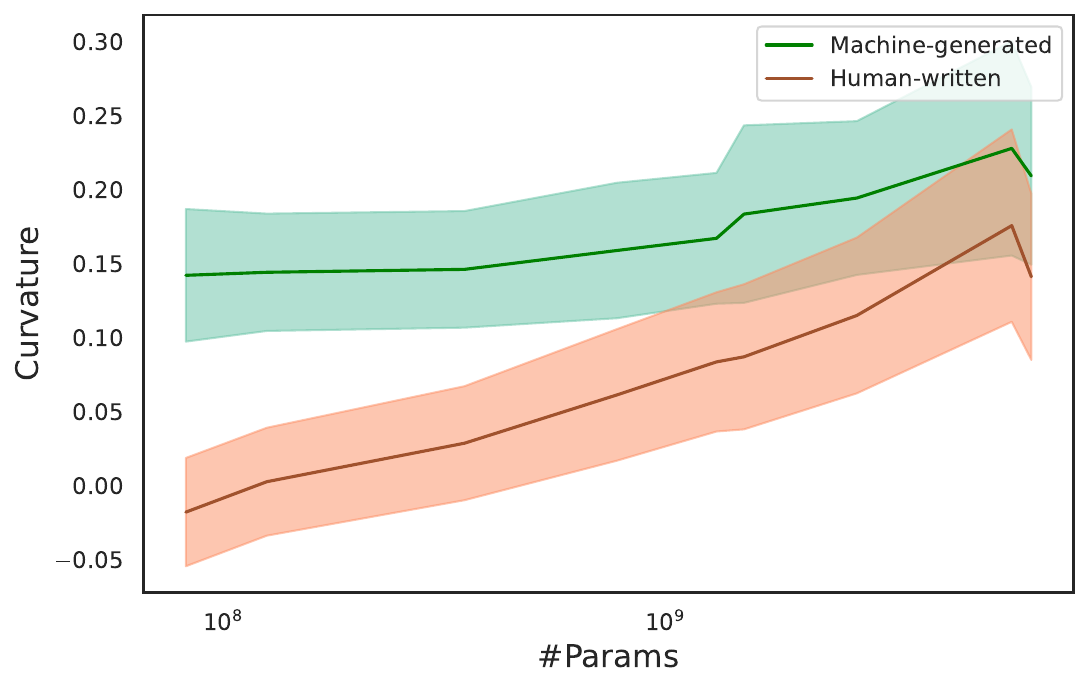}
     \footnotesize
     \caption{T5-Large}
     \label{fig:t5-large-dist}
    \end{subfigure}

    \begin{subfigure}{\linewidth}
        \centering
     \includegraphics[width=0.8\linewidth]{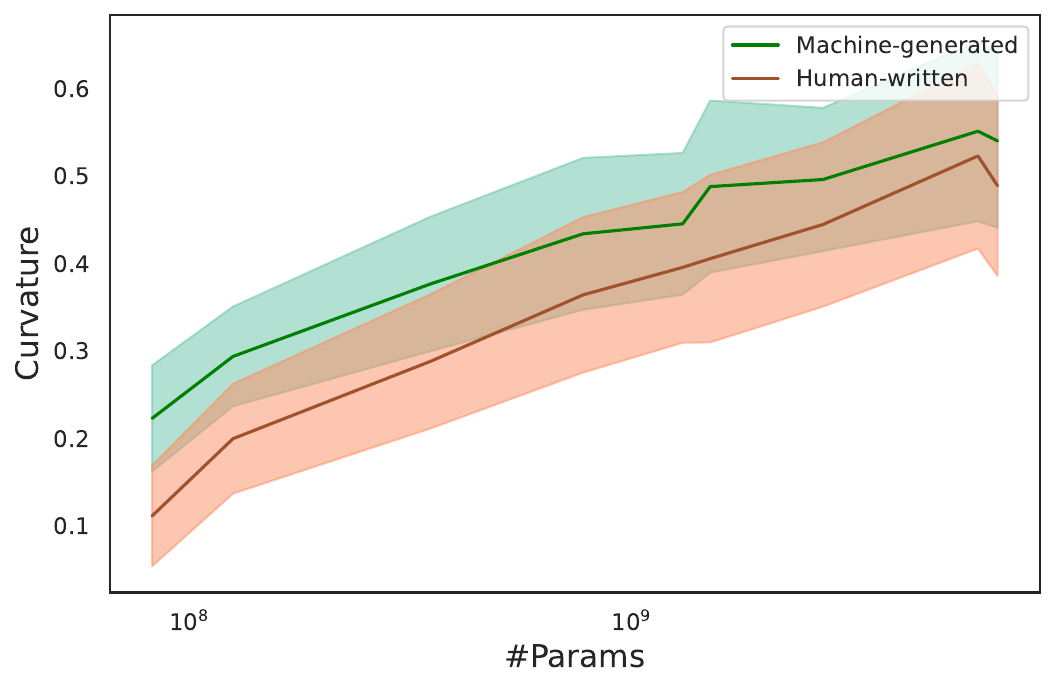}
     \footnotesize
     \caption{T5-Small}
     \label{fig:t5-small-dist}
    \end{subfigure}

    \begin{subfigure}{\linewidth}
     \centering
     \includegraphics[width=0.8\linewidth]{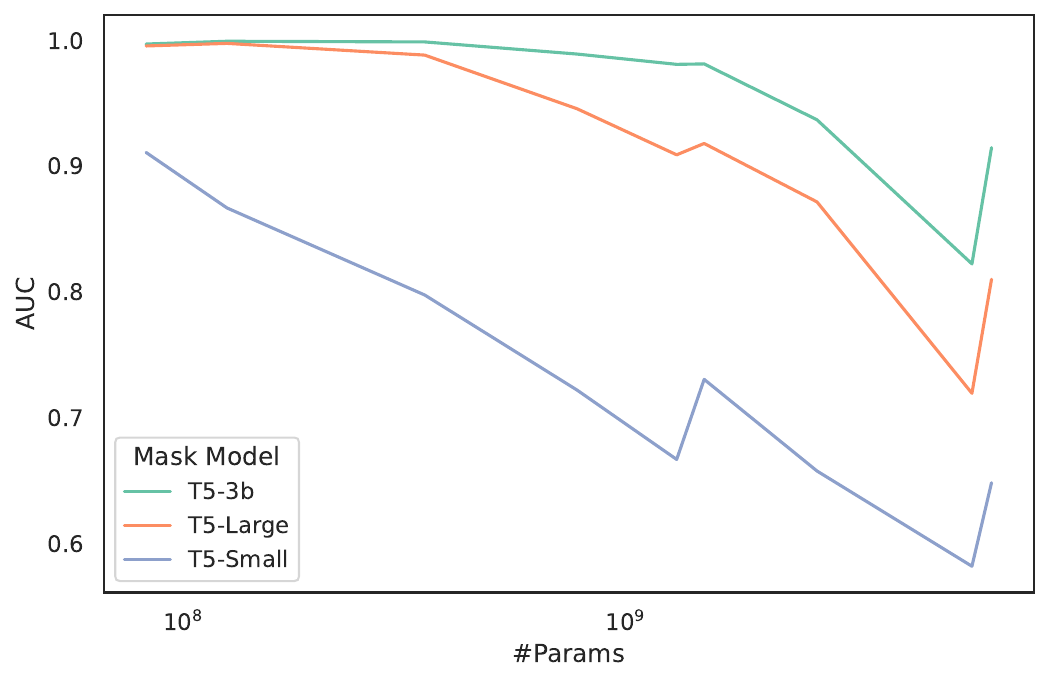}
     \footnotesize
     \caption{AUCs for different perturbation (masking)  models}
     \label{fig:t5-auc}
    \end{subfigure}
    
\caption{The effect of changing the perturbation (masking) model on curvature values and self-detection power of different models with different sizes (AUC).}
    \label{fig:t5-infill}
    \vspace{-2ex}
\end{figure}
%%%%%%%%%%%%%%%%%%%%%%%%%%%%%%%%%%%%

%\subsection{Mask Filling Model}

Figure~\ref{fig:t5-infill} shows the curvature numbers for each model trying to \textbf{detect its own} generations, so for each model the generator is also the detector. 
%In this experiment we want to see the effect of the mask filling model used for generating neighbors, on the estimated curvature value that is measured.
%
We experiment with three perturbation generating models, with three different sizes: (1) T5-small ($60$ million parameters) (2) T5-Large ($770$ million parameters) (3) T5-3B (3 billion parameter). The intuition behind using three model sizes is to see the effect of having a better replacement model on the measured curvatures and the detection power of the detector models.

We can see that as the masking model sizes down (going from top to the bottom subfigures), the overall curvature values for both human-written and machine-generated text increases (going from 0.2 maximum in Figure~\ref{fig:t5-3b-dist} to 0.6 maximum in Figure~\ref{fig:t5-small-dist}), and the two sets of texts become less distinguishable. 
T5-Small produces low-quality (low-fluency) neighbors that are assigned lower likelihoods by the detector model, resulting in high curvature numbers for both human and machine generated text, making them indistinguishable. As we improve the mask filling model, however, the generated neighbors become of higher quality (and semantically closer to the target point), thereby creating a more accurate estimate of the curvature and providing better distinguishablity, as shown by the AUC numbers in Figure~\ref{fig:t5-auc}.

\section{Experimental Setup}\label{app:exp_setup}

\subsection{Models}\label{sec:exp:models}
We want to experiment with a wide range of models, with different architectures, parameter counts and training datasets, therefore we use the following model families in our experiments: Facebook's OPT (we use the 125M, 350M, 1.3B, and 6.7B models), EleutherAI's GPT-J, GPTNeo and Pythia~\cite{biderman2023pythia} (we use GPTNeo-125M, GPTNeo-1.3B,  GPTNeo-2.7B, GPTJ-6B and Pythia models ranging from 70M to 2.8B parameters), and OpenAI's GPT models (distilGPT, GPT2-Small, GPT2-Medium, GPT2-Large, GPT2-XL, GPT-3 and ChatGPT). 

We also have experiments where we use partially trained models as detectors. For those experiments, we only use the Pythia models as they are the only ones with available, open-source partially trained checkpoints.
For each Pythia models, there is also a de-duplicated version available, where the model is trained on the de-duplicated version of the data, as opposed to the original dataset.
All the models we use are obtained from HuggingFace~\cite{hf}.

\subsection{Dataset}

\paragraph{Evaluation dataset.}
We follow~\citet{mitchell2023detectgpt}'s methodology for pre-processing and feeding the data. 
We use a subsample of the SQuAD dataset~\cite{rajpurkar2016squad}, where the original dataset sequences are used as the human-written text in the target sequence pool. We then use the first $20$ tokens of each human-written sequence as a prompt, and feed this to the target model, and have it generate completions for it. We then use this mix of generations and human-written text to create the target pool for which we do the detection.  
In all cases, following the methodology from~\citet{mitchell2023detectgpt}, our pool consists of $300$ human-written target samples, and $300$ machine-generated samples, so the overall pool size is $600$.

\paragraph{Pre-training datasets for the generative models.}
The {ElutherAI} and {Facebook} models (GPTJ, GPTNeo, Pythia and OPT families)  are all trained on the Pile dataset~\cite{gao2020pile}, a curated collection of $22$ English language datasets (consisting of web-crawled data, academic articles, dialogues, etc.). As mentioned above there are two versions of each Pythia model~\cite{biderman2023pythia}, one version is trained on Pile, the other is trained on de-duplicated Pile. 
The de-duplicated Pile is approximately 207B tokens in size, compared to the original Pile which contains 300B tokens.
There is limited information and access to the training data of the \textit{OpenAI} models. The GPT-2 family is reportedly trained on the WebText dataset, GPT-3 is trained on a combination of the Common Crawl~\footnote{\url{https://commoncrawl.org}}, WebText2, books and Wikipedia, and there is not any information released about the training data of ChatGPT.

% \subsection{Metrics}
% \paragraph{AUC}

\begin{figure}[t!]
\centering
      \includegraphics[width=0.9\linewidth]{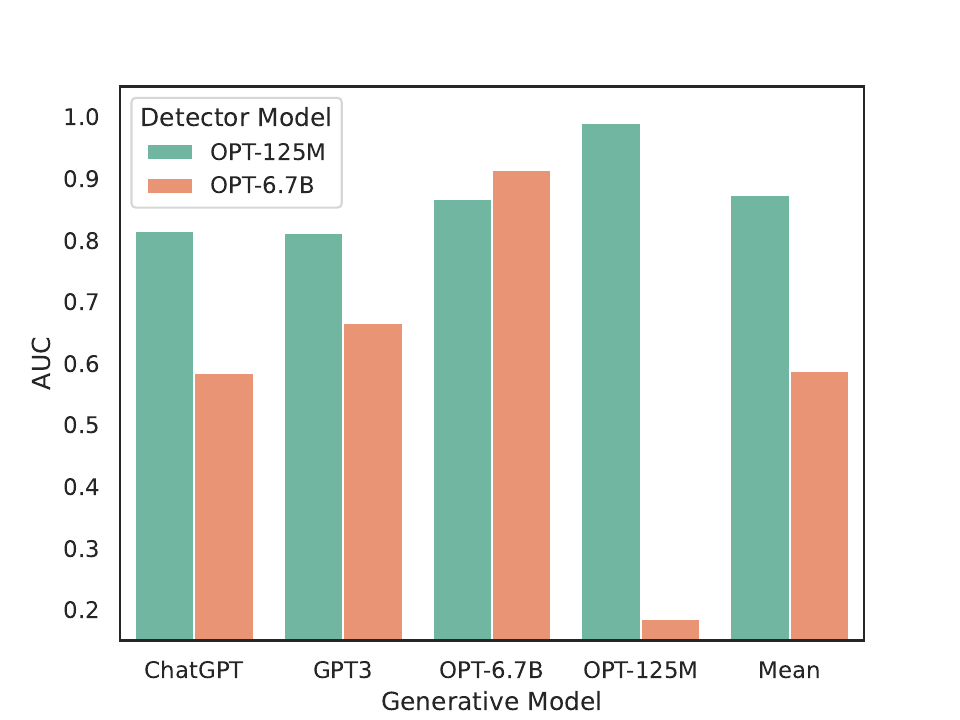}
     \caption{Summary of the cross-detection area under the ROC curve (AUC) results for a  selection of generative (the $4$ models over the X axis) and detector (OPT-125M and OPT-6.7B) models. We can see that the smaller OPT model is a better universal cross-detector. Full results are shown in Figure~\ref{fig:main_heatmap_size}.}  
       \label{fig:main_barplot_summary}
\end{figure}

\section{Additional Plots}

\subsection{Extensive Heatmaps}
We provide the full heatmaps from experiments of Section~\ref{sec:size} here, to provide a detailed breakdown.
Figures~\ref{fig:main_heatmap_size} and~\ref{fig:summary_heatmap_checkpoint} (full heatmap is Fig.~\ref{fig:main_heatmap_checkpoint} in Appendix) show the AUC of cross-detection for different models.
Figures~\ref{fig:main_heatmap_size_diff} and~\ref{fig:main_heatmap_checkpoint_diff} in Appendix show how close each detector comes, in terms of AUC, to self-detection.

\subsection{Summary of Experiments}

We provide a summary of Figure~\ref{fig:main_heatmap_size} in Figure~\ref{fig:main_barplot_summary}, where we have presented the numbers from the best overall detector with mean AUC of 0.92 (OPT-125M) and the biggest model of the same family, OPT-6.7B with average AUC of 0.46.

%This is an appendix.

\end{document}